\documentclass{jair}

\setcopyright{cc}
\acmDOI{10.1613/jair.1.xxxxx} 
\JAIRAE{Chang Xu}
\JAIRTrack{} 
\acmVolume{0}  
\acmArticle{0} 
\acmMonth{0}   
\acmYear{2026}

\RequirePackage[
  datamodel=acmdatamodel,
  style=acmauthoryear,
  backend=biber,
  giveninits=true,
  uniquename=init
  ]{biblatex}

\usepackage{subcaption}
\usepackage{pifont}
\usepackage{amsfonts}
\usepackage{amsmath}
\usepackage{tablefootnote}
\usepackage{multirow}
\usepackage{array}
\usepackage{makecell}

\graphicspath{ {./figures/} }

\usepackage{tikz}
\usetikzlibrary{positioning, shapes, arrows.meta}
\usepackage[edges]{forest}

\definecolor{hidden-draw}{RGB}{106,142,189}
\definecolor{hidden-blue}{RGB}{194,232,247}
\definecolor{hidden-orange}{RGB}{217, 232, 252}
\definecolor{hidden-red}{RGB}{255,182,193}
\definecolor{hidden-green}{RGB}{144,238,144}

\definecolor{cbYellow}{RGB}{240,228,66}
\definecolor{cbBlue}{RGB}{86,180,233}
\definecolor{cbOrange}{RGB}{230,159,0}
\definecolor{cbRed}{RGB}{213,94,0}
\definecolor{cbGreen}{RGB}{0,158,115}

\newcommand{\cmark}{\ding{51}}%

\begin{document}

\title[Towards Unified Approaches in Self-Supervised Event Stream Modeling]{Towards Unified Approaches in Self-Supervised Event Stream Modeling: Progress and Prospects}

\author{Levente Z\'{o}lyomi}
\authornote{Work conducted while the authors were with King AI Labs, Microsoft Gaming.}
\authornote{These authors contributed equally to this work.}
\email{levente.zolyomi@nx-ai.com}
\affiliation{%
  \institution{Johannes Kepler University}
  \city{Linz}
  \country{Austria}}
\affiliation{%
  \institution{NXAI GmbH}
  \city{Linz}
  \country{Austria}}

\author{Tianze Wang}
\authornotemark[1]\authornotemark[2]
\email{tianze.wang@kreditz.com}
\affiliation{%
  \institution{Kreditz AB}
  \city{Stockholm}
  \country{Sweden}}

\author{Sofiane Ennadir}
\email{sofiane.ennadir@king.com}
\affiliation{%
  \institution{King AI Labs, Microsoft Gaming}
  \city{Stockholm}
  \country{Sweden}}

\author{Oleg Smirnov}
\authornote{Corresponding author.}
\email{oleg.smirnov@microsoft.com}
\affiliation{%
  \institution{King AI Labs, Microsoft Gaming}
  \city{Stockholm}
  \country{Sweden}}

\author{Lele Cao}
\email{lele.cao@king.com}
\affiliation{%
  \institution{King AI Labs, Microsoft Gaming}
  \city{Stockholm}
  \country{Sweden}}

\renewcommand{\shortauthors}{Z\'{o}lyomi, Wang, Ennadir, Smirnov \& Cao}

\begin{abstract}
The proliferation of digital interactions across diverse domains, such as healthcare, e-commerce, gaming, and finance, has resulted in the generation of vast volumes of event stream (ES) data. ES data comprises continuous sequences of timestamped events that encapsulate detailed contextual information relevant to each domain. While ES data holds significant potential for extracting actionable insights and enhancing decision-making, its effective utilization is hindered by challenges such as the scarcity of labeled data and the fragmented nature of existing research efforts. Self-Supervised Learning (SSL) has emerged as a promising paradigm to address these challenges by enabling the extraction of meaningful representations from unlabeled ES data. In this survey, we systematically review and synthesize SSL methodologies tailored for ES modeling across multiple domains, bridging the gaps between domain-specific approaches that have traditionally operated in isolation. We present a comprehensive taxonomy of SSL techniques, encompassing both predictive and contrastive paradigms, and analyze their applicability and effectiveness within different application contexts. Furthermore, we identify critical gaps in current research and propose a future research agenda aimed at developing scalable, domain-agnostic SSL frameworks for ES modeling. By unifying disparate research efforts and highlighting cross-domain synergies, this survey aims to accelerate innovation, improve reproducibility, and expand the applicability of SSL to diverse real-world ES challenges.
\end{abstract}

\received{February 2025}
\received[accepted]{September 2026}

\maketitle

\section{Introduction}
\label{Introduction}

The rapid growth of digital interactions, ranging from online purchases and social media engagements to automated sensors in healthcare, continues to generate massive volumes of event stream (ES) data. Formally, an ES is a {\it continuous sequence of timestamped events, each encapsulating structured contextual information}. Industries such as healthcare, finance, gaming, and e-commerce have substantial ES repositories: healthcare systems store Electronic Health Records (EHRs) tracking patient admissions and diagnostic tests \cite{johnson2023mimic-iv,herrett2015data-cprd}, while e-commerce platforms log user interactions like clicks and purchases \cite{wang2019sequential,sun2019bert4rec}, among many other examples.

This surge of context-rich, time-dependent ES data holds potential for a range of downstream applications: e.g., diagnosing a patient’s evolving condition in healthcare \cite{li2020behrt}, detecting fraudulent transactions in finance \cite{babaev2022coles}, or personalizing game content to enhance player engagement \cite{wang2024player2vec,pu2022unsupervised}. However, large-scale event datasets typically lack the extensive {\it labeling} needed to train traditional supervised learning systems. Moreover, existing modeling efforts often remain {\it fragmented} across domains, sometimes replicating research efforts without leveraging the many structural similarities shared by event streams across industries. These challenges underscore the key demand for ES modeling approaches -- ones that can meaningfully learn from raw event logs with minimal or no human annotation.

\subsection{Aim and Scope}

Self-Supervised Learning (SSL) stands out as a powerful paradigm well-suited to address these challenges. By deriving supervisory signals directly from unlabeled data, SSL can learn informative representations of sequences without relying on manual annotation. In the context of ES, SSL enables models to capture temporal dynamics and contextual relationships, which are often critical for tasks such as predicting future events, clustering entity behaviors, or detecting anomalies. Despite the evidence that SSL has shown major advances in fields like language modeling~\cite{devlin2018bert,brown2020language} and computer vision~\cite{he2022masked,oquab2023dinov2}, its application to ES modeling has not yet converged into a unified body of knowledge.
Current ES research typically targets a specific domain (e.g., healthcare, gaming, e-commerce, finance) without systematically building on or comparing methods across these verticals.

Therefore, the purpose of this survey is to (1) unify the progress in SSL for event streams across multiple domains, (2) identify critical gaps and challenges that cut across those fields, and (3) propose future directions to foster development of more general, domain-agnostic SSL approaches for ES modeling. By addressing these diverse aspects, we aim to accelerate innovation, improve reproducibility, and highlight emerging themes that can benefit the ES modeling community at large. Controlled comparisons of SSL paradigms under matched conditions are still lacking, so the synthesis and guidance we offer rest on observed patterns in the literature rather than systematic empirical evidence (Sections~\ref{subsec:ssl-summary} and~\ref{subsec:downstream-tsks}).

\begin{figure}[ht!]
    \centering
    \begin{subfigure}[b]{0.48\textwidth}
        \centering
        \includegraphics[width=\textwidth]{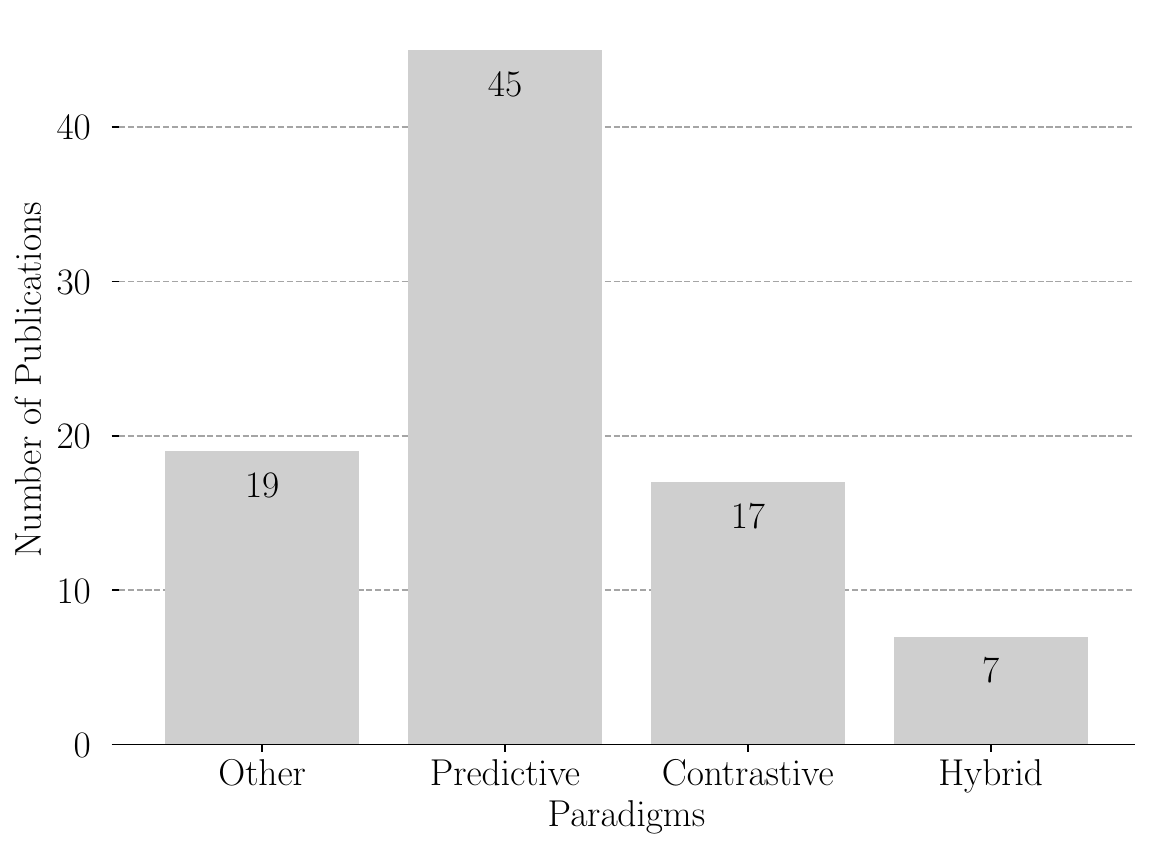}
        \caption{}
        \label{fig:literature-ssl-stats}
    \end{subfigure}
    \begin{subfigure}[b]{0.48\textwidth}
        \centering
        \includegraphics[width=\textwidth]{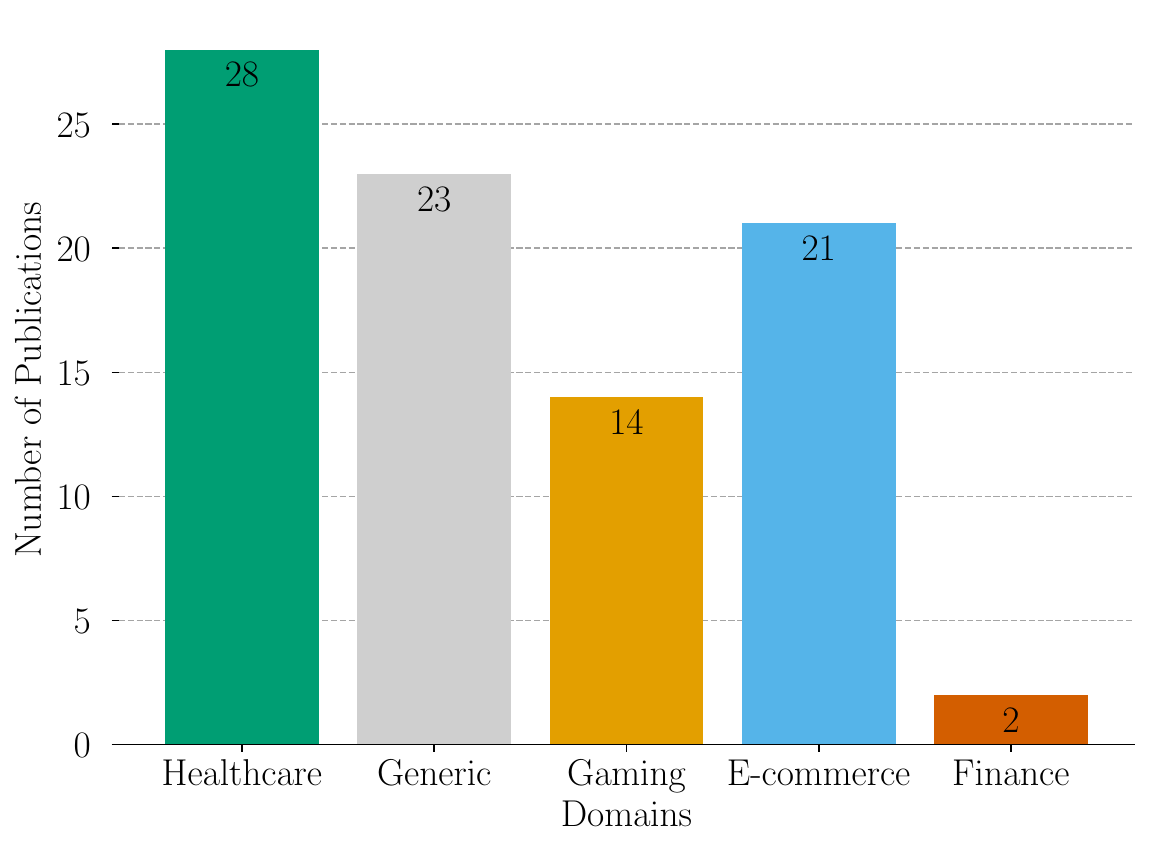}
        \caption{}
        \label{fig:literature-domain-stats}
    \end{subfigure}

    \caption{Distribution of the reviewed works across (a) various SSL paradigms and (b) diverse application domains.}
    \Description{Two bar charts. The left chart counts reviewed papers per SSL paradigm, with predictive paradigms such as masked modeling, autoregressive modeling, and temporal point processes accounting for most works and contrastive paradigms for fewer. The right chart counts reviewed papers per application domain, covering healthcare, e-commerce, finance, gaming, and domain-generic methods.}
    \label{fig:literature-statistics}
\end{figure}

To ensure comprehensive yet focused coverage, we reviewed papers from leading conferences such as ICML, NeurIPS, AAAI, and KDD, as well as journals in machine learning, artificial intelligence, and knowledge discovery, along with domain-specific journals, such as those in healthcare. Our search queries combined SSL-related terms (e.g., ``self-supervised learning'', ``pre-training'', ``contrastive learning'', ``masked modeling'') with ES-related terms (e.g., ``event sequences'', ``event log'', ``electronic (health) records'', ``behavior data''). We included papers that (1) address sequential or temporal event data as defined in Section~\ref{subsec:def-notation}, and (2) employ self-supervised pre-training objectives and related methodologies. Beginning with an initial pool of $\approx 100$ works, we refined our selection by prioritizing recent studies introducing novel concepts alongside seminal works with substantial impact on the field, as evidenced by their citation influence. The final distribution of surveyed works is presented in Figure~\ref{fig:literature-statistics}. The comparatively small number of finance works reflects the current state of the published literature, where event-level financial data is typically proprietary (Section~\ref{subsec:datasets-overview}), rather than a scoping choice of this survey. Our temporal scope begins around 2016, coinciding with the introduction of neural architectures for temporal point processes \cite{du2016recurrent} and the broader adoption of deep learning for sequential data. Earlier work on event sequences relied predominantly on classical statistical methods and handcrafted features, which fall outside the scope of modern SSL paradigms that form the focus of this survey. Compared with such classical approaches, SSL is most beneficial when labeled data is scarce relative to the volume of raw event logs, when representations must transfer across multiple downstream tasks, or when event attributes are too heterogeneous for handcrafted features. For small, well-labeled, single-task problems, classical supervised baselines can remain competitive. Overall, this approach ensures our review captures both the breadth and depth of SSL methods applied to ES data.

\begin{table}[ht]
    \caption{Comparison of related surveys' coverage for modalities and application domains.}
    \label{tab:survey-comparison}
    \centering
    \scriptsize
    \renewcommand{\arraystretch}{1.8} 
    \setlength{\tabcolsep}{4pt} 
    \begin{tabular}{c|c|c|c|c|c|c|c|c|c|c|c}
        &
        & \rotatebox{75}{\citet{shchur2021neural}}
        & \rotatebox{75}{\citet{shickel2017deep}}
        & \rotatebox{75}{\citet{dong2022table}}
        & \rotatebox{75}{\citet{zhang2024self}}
        & \rotatebox{75}{\citet{wornow2023shaky}}
        & \rotatebox{75}{\shortstack{\citet{wang2019sequential}\\\citet{wang2021survey}}}
        & \rotatebox{75}{\citet{yu2023self}}
        & \rotatebox{75}{\citet{hooshyar2018data}}
        & \rotatebox{75}{\citet{liang2024foundation}}
        & \rotatebox{75}{\textbf{Our work}} \\ \hline

        \multirow{3}{*}{\rotatebox{90}{\textbf{Modality}}}
        & \makecell{Continuous-Time\\ Event Sequences}
        & \cmark &  &  &  & \cmark & \cmark & \cmark &  &  & \cmark \\ \cline{2-12}

        & Tabular
        &  &  \cmark & \cmark &  &  &  &  & \cmark &  & \cmark \\ \cline{2-12}

        & Time Series
        &  &  &  & \cmark &  &  &  &  & \cmark & \cmark \\ \hline

        \multirow{4}{*}{\rotatebox{90}{\textbf{Domain}}}
        & Healthcare
        & \cmark & \cmark &  & \cmark & \cmark &  &  &  & \cmark & \cmark \\ \cline{2-12}

        & Finance
        &  &  &  & \cmark &  &  &  &  & \cmark & \cmark \\ \cline{2-12}

        & Gaming
        &  &  &  &  &  &  &  & \cmark &  & \cmark \\ \cline{2-12}

        & E-commerce
        & \cmark &  &  &  &  & \cmark & \cmark &  &  & \cmark
    \end{tabular}
\end{table}

\subsection{Contribution and Structure}
This survey offers several key contributions:
\begin{itemize}
    \item {\bf Comprehensive synthesis across domains}: We provide the first (to our knowledge) cross-domain assessment of SSL-based ES modeling. While prior work has surveyed either specific application domains (e.g., healthcare, e-commerce, and gaming) or other data modalities (e.g., tabular, time series, and continuous-time sequences), we argue that ES data has unifying structural properties that make it a distinct category meriting an overarching view. We consider ES as a common structure across domains, grounded in the unified formal framework of Section~\ref{subsec:def-notation}, and organize SSL methods according to their learning objectives rather than their application context. Table~\ref{tab:survey-comparison} situates this perspective relative to existing surveys, highlighting that our focus is on synthesizing shared methodological patterns across previously siloed domains. 

    \item {\bf Taxonomy of SSL paradigms for ES}: We present a structured taxonomy that organizes the wide spectrum of SSL methods used in ES modeling, ranging from {\it predictive} techniques (e.g., masked modeling, autoregressive strategies, and temporal point processes) to various {\it contrastive} approaches (e.g., instance-based, distillation-based, and multimodal). By emphasizing shared technical motifs, we identify cross-domain synergies that can strengthen and unify existing approaches, and we map each paradigm to the downstream task categories it best serves (Sections~\ref{subsec:ssl-summary} and~\ref{subsec:downstream-tsks}).
    \item {\bf Resource and benchmark guide}: We compile an overview of widely used public datasets, identify limitations in current benchmarking practices, and suggest how future research can more reliably evaluate and compare models. We also point out the scarcity of multi-domain benchmarks for ES and the need for stronger open-source resources.
    \item {\bf Roadmap for future research}: We outline critical open issues in ES SSL, ranging from timestamp modeling to more robust augmentations, and discuss how next-generation methods could push the boundaries of performance, scalability, and interpretability in real-world applications.
\end{itemize}

The remainder of this paper is organized as follows. Section~\ref{sec:event-stream-data} introduces the fundamentals of event streams and key application domains. Section~\ref{sec:ssl-es-review} discusses SSL paradigms, including predictive and contrastive approaches, and presents our taxonomy of existing SSL-based methods for event streams. Section~\ref{sec:es-data-overview} reviews common ES datasets and downstream evaluation tasks. In Section~\ref{sec:progress-prospects}, we synthesize open challenges and propose future research directions. Finally, Section~\ref{sec:conclusion} concludes with key findings and broader implications for data-driven ES modeling.

\section{Foundations of Event Streams}
\label{sec:event-stream-data}

Event Stream (ES) data provides a flexible and expressive data format for representing and modeling sequential events across diverse domains. In the following subsections, we (1) establish the definition and notation (Section~\ref{subsec:def-notation}) for the structured foundation and key components for reasoning about ES data, (2) introduce the relation of ES data to other data modalities (Section~\ref{subsec:other-modality}), and (3) highlight the application domains and potential challenges  (Section~\ref{subsec:application-domains}) when working with ES data.

\subsection{Definition and Notation}
\label{subsec:def-notation}

An ES is a continuous, ordered sequence of events generated over time by one or more sources. Each event encapsulates timestamped information about a specific action or state. While each event is a discrete occurrence, the ES represents an ongoing, continuous flow of events, which is a common characteristic of real-world scenarios.

An \textit{entity} $u$, such as a user or a patient, is typically associated with each ES. Formally, an ES for an entity $u$ can be defined as a sequence:
\begin{equation}
\label{eq:es-def}
S_u = \{e_{u,i}\}_{i=1}^{\infty}, \text{where } e_{u,i} = (t_{u,i}, d_{u,i}).
\end{equation}
$e_{u,i}$ represents the $i$-th event in the ES for entity $u$. $t_{u,i} \in \mathcal{T}$ is the timestamp of $e_{u,i}$ from a totally ordered time domain $\mathcal{T}$, ensuring $t_{u,i} \leq t_{u,i+1}$ for all $i \geq 1$. $d_{u,i} \in \mathcal{D}$ represents all information associated with $e_{u,i}$, where $\mathcal{D}$ is the domain of all possible event data.

All observed entities in a dataset are represented as $U = \{u_i\}_{i=1}^m$, where $m$ denotes the total number of entities. To simplify notation, the entity indicator $u$ is omitted in sequences when it is contextually clear that the events belong to the same entity. The set of all sequences across all entities is represented as an ES dataset, $\mathcal{S} = \{S_1, S_2, \ldots, S_m\}$.

\begin{figure}[ht]
    \centering
    \includegraphics[width=0.8\linewidth]{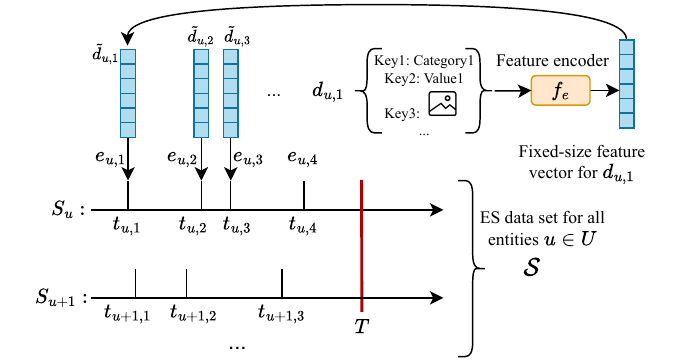}
    \caption{Illustration of ES data for individual entities $u, u+1, \ldots$. Each event in an entity’s ES, as exemplified using $e_{u,1}$, is characterized by a timestamp and associated data, which is encoded into a fixed-size feature vector $\Tilde{d}_
    {u,1}$ using a feature encoder $f_e$. While the encoder output could in principle be matrix-valued, pooled vector representations are used in practice as downstream models typically operate on vector features. The diagram showcases the temporal sequence of events for multiple entities and highlights the role of the encoder in transforming diverse event data into standardized representations for analysis.}
    \Description{Horizontal timelines for several entities, each carrying discrete events at irregular time points up to an observation time marked by a vertical line. A zoomed view of one event shows its timestamp and raw attributes, such as categorical and numerical fields, passing through a feature encoder to produce a fixed-size embedding vector.}
    \label{fig:es-data-vis}
\end{figure}

Although an ES is theoretically infinite in length as new events continuously occur, practical analysis often considers a finite realization of the stream up to a specific time $T \in \mathcal{T}$. This realization contains only the events observed until time $T$, as indicated by the red vertical line in Figure~\ref{fig:es-data-vis}. Notably, event streams from different entities, such as $u_1$ and $u_2$, may exhibit varying numbers of events recorded within the same time frame up to $T$.

In its most common form, $d_i$ represents data from various modalities, often structured as a dictionary containing an arbitrary number of key-value pairs. For simplicity, we omit the entity indicator $u$ and refer to $d_i$ instead of $d_{u,i}$. For example, in the context of a patient undergoing an X-ray, the event data might include the X-ray image, the hospital's identifier, details about the imaging equipment, and the physician's evaluation.

In practice, the information encapsulated in $d_i$ is typically compressed into a fixed-length $n$-dimensional vector representation, denoted as $f_e(d_i) = \Tilde{d}_i \in \mathbb{R}^n$, via an event data encoder $f_e: \mathcal{D} \rightarrow \mathbb{R}^n$, where $n$ represents the embedding size for all event representations $\Tilde{d}_i$. However, recent advances in multimodal learning suggest that richer representations, such as multiple vectors or matrix-valued encodings, may better preserve the internal structure of complex events (e.g., medical images with accompanying metadata, or product listings with visual and textual attributes). Exploring such extensions remains an open direction, though it falls beyond the scope of this survey. It is also worth noting that designing an effective encoder for diverse event data types is a complex and nontrivial task in its own right. A visual depiction of this representation is provided in Figure~\ref{fig:es-data-vis}.

\subsection{Relation to Other Modalities}
\label{subsec:other-modality}
ES data can be transformed into various modalities to leverage established machine learning models and techniques. Each of these transformations facilitates the use of methods from well-studied modalities but introduces specific trade-offs, such as sparsity in tabular formats or limitations on event data space in time series representations. For instance, one approach is to convert ES data into a tabular format where each row represents an event, with columns capturing its associated attributes~\cite{shickel2017deep}. However, this approach often results in sparse datasets with numerous missing or irrelevant entries, which can adversely affect model performance~\cite{wang2020mimic-extract}. Furthermore, tabular representations fail to incorporate the temporal dimension, making it difficult to capture the evolving dynamics inherent in ES data.

Another approach is to transform ES data into a time series, treating events as a continuous sequence. While this transformation can model temporal trends, it assumes that data is sampled at regular intervals, which is often not the case for ES, where events occur irregularly over time~\cite{shukla2021multi}. Additionally, this method restricts the data space $\mathcal{D}$ by constraining events to time-indexed continuous values, excluding categorical, ordinal, and other discrete outcomes~\cite{mcdermott2021comprehensive}.

Dynamic graph representations offer another approach for converting ES data, where entities are modeled as nodes and events or interactions as time-evolving edges~\cite{kazemi2020representation}. While graphs excel at capturing relational structures, applying them to ES data can introduce challenges. The inherent temporal granularity of ES may result in highly dynamic graphs with frequent edge updates, leading to computational inefficiencies and scalability issues~\cite{skarding2021foundations}. Furthermore, the evolving nature of dynamic graphs can obscure the sequential order of events, which is crucial for many downstream tasks.

Moreover, the definition of ES extends the framework of modeling continuous-time event sequences with temporal point processes (TPPs)~\cite{du2016recurrent}. While TPPs focus on generative mechanisms for event occurrence with an emphasis on event timing, they do not address the broader objectives of self-supervised ES modeling and downstream tasks. TPPs are often constrained by specific assumptions about event data, such as discrete markers in marked TPPs~\cite{mei2017neural} or continuous spatial domains in spatiotemporal point processes~\cite{chen2020neural}. These limitations reduce the ability to incorporate diverse metadata associated with real-world events and restrict the flexibility of TPPs in learning versatile entity representations for downstream applications. ES data modeling overcomes these constraints by supporting a wider range of event types and flexible learning objectives, which are critical for diverse applications of SSL.

\subsection{Application Domains and Challenges}
\label{subsec:application-domains}
Throughout this survey, we use \emph{domain} to denote broad application verticals (e.g., healthcare, e-commerce) that generate ES through entity-system interactions. This notion is distinct from data modality (e.g., images, text) or downstream task. We focus on domains where (1) ES are the primary carrier of behavioral information and (2) a substantive SSL literature already exists. Other settings, including DevOps logs~\cite{xie2024logsd}, cyber-security streams~\cite{wu2025robust}, and service telemetry, share ES structural properties but lack mature SSL methods; we note these as emerging opportunities rather than surveyed domains. Conversely, video, while representable as sequences of image-valued events, is better treated within visual time-series research traditions.

Below, we describe four widely studied domains that illustrate both the diversity of ES applications and the shared challenges motivating domain-agnostic SSL methods. Each domain presents distinct data characteristics and modeling requirements, yet all exhibit structural similarities that make cross-domain transfer a compelling research direction. A mapping of ES concepts across these domains appears in Table~\ref{tab:notation}.

\begin{table}[ht!]
    \caption{Mapping of ES concepts to examples across different application domains.}
    \label{tab:notation}
    \centering
    \footnotesize
    \renewcommand{\arraystretch}{1.5} 
    \setlength{\tabcolsep}{2pt} 
    \begin{tabular}{>{\centering\arraybackslash}p{0.15\linewidth}|
                        >{\centering\arraybackslash}p{0.2\linewidth}|
                        >{\centering\arraybackslash}p{0.2\linewidth}|
                        >{\centering\arraybackslash}p{0.2\linewidth}|
                        >{\centering\arraybackslash}p{0.2\linewidth}}
        \textbf{Concept} & \textbf{Healthcare} & \textbf{Finance} & \textbf{Gaming} & \textbf{E-commerce} \\ \hline
        Entity & Patient & Stock ticker or account & Player & Customer \\ \hline
        Event & Hospital admission, diagnostic test & Buy/sell order, transaction log & Player action, game state change & Page view, click, purchase \\ \hline
        Categorical Feature & Hospital department, diagnosis type & Transaction type (buy/sell), stock category & Item type, game level & Product category, payment method \\ \hline
        Numerical Feature & Patient's age, test result & Transaction amount, stock price & In-game score, player level & Item price, time spent on page \\ \hline
        Timestamp & Time of admission, test completion & Time of transaction, stock price update & Time of action, level completion & Time of click, purchase \\ \hline
        Event Stream & Sequence of medical events in an EHR & Sequence of trades or transactions for a stock or account & Sequence of player actions or game events & Sequence of user interactions (clicks, purchases, etc.)
    \end{tabular}
\end{table}

\begin{enumerate}
\item \textbf{Healthcare}: Hospitals and clinics record a continuous flow of patient information, such as admissions, diagnoses, lab results, and treatments. These clinical events, each characterized by a timestamp and metadata, form a patient’s EHR. Modeling patient trajectories through EHR data can unlock valuable insights for disease progression, risk stratification, and personalized care ~\cite{johnson2023mimic-iv,herrett2015data-cprd}. However, privacy regulations (e.g., HIPAA\footnote{\url{https://www.hhs.gov/hipaa}} and GDPR\footnote{\url{https://gdpr-info.eu}}) limit data sharing, hindering the development of large-scale public benchmarks. Moreover, healthcare data often exhibits noise from measurement errors, missing event attributes, and inconsistent documentation practices across different medical institutions.

\item \textbf{E-commerce}: User interactions, such as product page views, clicks, cart additions, and purchases, are recorded as event streams on online platforms. Sequential recommender systems (SRS)~\cite{sun2019bert4rec,zhou2020s3} rely on these data to predict future user actions and personalize the shopping experience. While typically abundant, e-commerce event streams can suffer from {\it implicit feedback} noise (e.g., accidental clicks or browsing behavior not directly correlated with user interest). Additionally, item catalogs and user demographics change over time, introducing concept drift that demands adaptive modeling techniques.

\item \textbf{Finance}: Financial institutions and trading platforms generate event streams through a variety of transactions, including stock trades, credit card purchases, and blockchain-based transfers~\cite{hu2023bert4eth,babaev2022coles}. In these contexts, timeliness and accuracy are paramount, as decisions often depend on real-time signals for fraud detection or high-frequency trading. Financial ES data may involve incomplete transaction details or delayed reporting, complicating model training and deployment. Moreover, financial data often contains sensitive or proprietary information, leading to restricted availability of large-scale public datasets.

\item \textbf{Gaming}: Game developers log in-game actions such as user movements, item pickups, and interactions with non-player characters. This data helps to analyze gameplay, improve design, and personalize experiences~\cite{kantharaju2020discovering,min2016player}. However, gaming logs can be extremely noisy -- players may experiment randomly, idle in the game, or perform repetitive tasks to ``farm'' resources. The rapid and often unpredictable evolution of game rules, patches, or newly introduced features further exacerbates modeling difficulties. Additionally, balancing real-time feedback (e.g., matchmaking) with large-scale event processing remains a significant engineering challenge.
\end{enumerate}

While the details differ by application domain, ES data in these domains shares common difficulties: (1) \textit{irregular sampling}, as event arrivals can be sporadic, requiring flexible timestamp modeling; (2) \textit{diverse feature types}, because each event can encompass a wide range of features, from discrete labels to unstructured text or images; and (3) \textit{privacy and ethical concerns}, especially in healthcare or finance, where the improper handling of sensitive records has serious legal and ethical implications. These overlapping challenges collectively motivate the exploration of self-supervised approaches that may effectively leverage unlabeled event streams, mitigate noise, and handle complex temporal dependencies across domains. In highly regulated settings, centralized data aggregation is often infeasible, motivating federated or privacy-preserving approaches that train models across institutions without exposing raw event logs~\cite{rieke2020future}.

\section{Self-Supervised Learning for Event Streams}
\label{sec:ssl-es-review}
Self-supervised learning has become a cornerstone for ES modeling, as it enables the effective utilization of vast volumes of unlabeled data -- a common characteristic of real-world ES datasets. By designing alternative tasks, known as ``pretext'' tasks, SSL uses the data itself as supervisory signals~\cite{liu2021self}. This approach is particularly crucial for ES modeling, where labeled data is often sparse, incomplete, or expensive to obtain. Through pretext tasks, SSL facilitates the extraction of meaningful representations during an initial pre-training phase, capturing the complex temporal and contextual relationships inherent in ES data. These pre-trained representations serve as a robust foundation, allowing models to either be fine-tuned on limited labeled datasets for specific downstream tasks or directly applied in a zero-shot manner, bypassing the need for additional task-specific training. This dual capability makes SSL not only instrumental but often a prerequisite for building effective ES modeling pipelines in diverse domains.

\begin{figure}[ht!]
    \centering
    \forestset{
  leaf/.style={},
  framework/.style={draw=darkgray, dashed, line width=1pt},
  lossvariant/.style={fill=gray!8},
  datasetting/.style={fill=gray!8},
  healthcare/.style={fill=cbGreen!20},
  ecommerce/.style={fill=cbBlue!20},
  finance/.style={fill=cbRed!20},
  gaming/.style={fill=cbOrange!20}
}
\begin{forest}
forked edges,
for tree={
    grow=east,
    reversed=true,
    anchor=base west,
    parent anchor=east,
    child anchor=west,
    node options={align=center},
    align = center,
    base=left,
    font=\small,
    rectangle,
    draw=hidden-draw,
    rounded corners,
    minimum width=4em,
    edge+={darkgray, line width=1pt},
    s sep=3pt,
    inner xsep=2pt,
    inner ysep=3pt,
    ver/.style={rotate=90, child anchor=north, parent anchor=south, anchor=center},
},
where level=1{text width=4em,font=\scriptsize}{},
where level=2{text width=5em,font=\scriptsize}{},
where level=3{text width=4.8em,font=\scriptsize}{},
where level=4{text width=4.8em,font=\scriptsize}{},
[
Self-Supervised Learning, ver
[
Contrastive\\SSL
[
Instance\\Contrastive
[
Generic
[
\citet{xue2022hypro,wang2023hierarchical}, leaf, text width=18em, align=left
]
]
[
Healthcare
[
\citet{raghu2023sequential}, leaf, healthcare, text width=18em, align = left
]
]
[
E-commerce
[
\citet{ma2020disentangled,xie2022contrastive} \\
\citet{hou2022towards,qiu2022contrastive} \\
\citet{wang2023sequential}, leaf, ecommerce, text width=18em, align = left
]
]
[
Finance
[
\citet{babaev2022coles}, leaf, finance, text width=18em, align = left
]
]
]
[
Distillation, lossvariant
[
Healthcare
[
\citet{li2022hi-behrt}, leaf, healthcare, text width=18em, align = left
]
]
]
[
Feature \\ Decorrelation, lossvariant
[
Healthcare
[
\citet{raghu2023sequential}, leaf, healthcare, text width=18em, align = left
]
]
]
[
Multimodal\\Contrastive,
[
Healthcare
[
\citet{king2023multimodal,ma2024global}
, leaf, healthcare, text width=18em, align = left
]
]
]
]
[
Predictive\\SSL
[
Masked\\Modeling
[
Healthcare
[
\citet{li2020behrt,kodialam2021deep} \\
\citet{meng2021bidirectional,rasmy2021med} \\
\citet{labach2023duett,yang2023transformehr}
, leaf, healthcare, text width=18em, align = left
]
]
[
E-commerce
[
\citet{sun2019bert4rec,zhou2020s3}
, leaf, ecommerce, text width=18em, align = left
]
]
[
Finance
[
\citet{hu2023bert4eth}, leaf, finance, text width=18em, align = left
]
]
[
Gaming
[
\citet{pu2022unsupervised,wang2024player2vec}, leaf, gaming, text width=18em, align = left
]
]
]
[
Autoregressive\\Modeling
[
E-commerce
[
\citet{tang2018personalized,li2018learning} \\
\citet{yuan2019simple,ying2018sequential} \\
\citet{li2020time,ma2020disentangled} \\
\citet{wang2020jointly,xie2022contrastive} \\
\citet{hou2022towards,wang2023sequential} \\
\citet{luo2023mcm}, leaf, ecommerce, text width=18em, align = left
]
]
[
Gaming
[
\citet{min2016player} \\
\citet{kantharaju2020discovering} \\
\citet{zhao2021player}, leaf, gaming, text width=18em, align = left
]
]
]
[
Temporal\\ Point\\ Processes, framework
[
Generic
[
\citet{du2016recurrent,mei2017neural}\\
\citet{guo2018initiator,omi2019fully}\\
\citet{shchur2019intensity,mei2020noise}\\
\citet{zhang2020self-attentive,zuo2020transformer-hawkes}\\
\citet{chen2020neural,xue2022hypro}\\
\citet{yang2022transformer,wang2023hierarchical}\\
\citet{zhang2024neural}, leaf, text width=18em, align = left
]
]
[
Healthcare
[
\citet{enguehard2020neural,steinberg2023motor}, leaf, healthcare, text width=18em, align=left
]
]
]
]
]
]
\end{forest}
    \caption{Overview of SSL methods for ES in application to \textcolor{cbBlue}{e-commerce}, \textcolor{cbGreen}{healthcare}, \textcolor{cbOrange}{gaming}, and \textcolor{cbRed}{finance}, as well as generic methods that are not tailored to any specific domain. Solid borders denote core paradigms; dashed borders indicate a generative modeling framework included for its predictive character; shaded nodes denote loss function variants.}
    \Description{A tree diagram rooted at SSL for event streams with two main branches. The predictive branch splits into masked modeling, autoregressive modeling, and temporal point processes; the contrastive branch splits into instance contrastive, distillation, feature decorrelation, and multimodal contrastive. Leaves list representative methods on light background tints that encode the application domain.}
    \label{fig:ssles-tree-diagram}
\end{figure}

Our taxonomy reflects how the ES literature is currently organized rather than a theoretically pure decomposition into orthogonal design axes. This means our categories span different conceptual levels: masked and autoregressive modeling denote training objectives; distillation and feature decorrelation are loss function variants; multimodal contrastive addresses a specific data setting; and temporal point processes constitute a generative modeling framework included for its predictive character and practical importance. We adopt this structure because it mirrors the decisions practitioners actually face: researchers typically choose a high-level paradigm (predictive vs. contrastive), then select among established variants within that paradigm, rather than independently combining objectives, losses, and data assumptions from first principles. Throughout the section, we note where categories are orthogonal and could be combined; such unexplored combinations represent opportunities for future work.

In the following subsections, we first give a brief overview of SSL principles. We then organize current ES research into two major families: (1) \emph{predictive SSL}, which learns by predicting masked, future, or temporally structured aspects of the event stream (Section~\ref{subsec:predictive-ssl}), and (2) \emph{contrastive SSL}, which learns a representation space that brings similar event sequences closer and pushes dissimilar ones apart (Section~\ref{subsec:contrastive-ssl}). This taxonomy is visualized in Figure~\ref{fig:ssles-tree-diagram}, which also shows how these paradigms manifest across the domains specified in Table~\ref{tab:notation}. Formal definitions of each objective appear in Appendix~\ref{appendix:ssl-paradigms}.

\subsection{Overview of SSL Principles}
\label{subsec:ssl-overview}

In essence, SSL aims to create {\it pretext tasks} that derive supervisory signals directly from unlabeled data. For event streams, such tasks typically involve:
\begin{itemize}
\item \emph{Masking} or \emph{corrupting} certain events or attributes, then training the model to reconstruct them.
\item \emph{Autoregressive} next-event (or next-token) prediction, relying only on previously observed events.
\item \emph{Comparing} sequences or subsequences under various augmentations to learn semantic representations.
\end{itemize}

Because ES data often contains rich metadata (Section~\ref{subsec:def-notation}) with timestamped and potentially high-dimensional event features, SSL methods can exploit temporal structure, complex contextual cues, and large-scale unlabeled sequences for representation learning. This stands in contrast to supervised learning, which is limited by the availability of high-quality labels in each domain. SSL, therefore, accelerates progress by leveraging all observed events, including those without explicit annotation, to learn robust event representations, enabling further fine-tuning or zero-shot applications for downstream tasks.

Standard SSL paradigms in broader scope include \emph{predictive} approaches such as Masked Vision Autoencoders~\cite{he2022masked} and BERT~\cite{devlin2018bert}, along with \emph{contrastive} methods exemplified by SimCLR~\cite{chen2020simple} and MoCo~\cite{he2020momentum}. As we show in Sections~\ref{subsec:predictive-ssl} and \ref{subsec:contrastive-ssl}, both of these map naturally onto ES data, while also accommodating the unique constraints posed by irregular sampling or multi-modal event attributes.

\subsection{Predictive SSL}
\label{subsec:predictive-ssl}

Predictive SSL focuses on learning representations by predicting or reconstructing masked or missing parts of the event stream. This family of methods often draws inspiration from natural language processing (NLP), where large language models (LLMs) are pre-trained to predict masked tokens such as BERT \cite{devlin2018bert} or the next token in an autoregressive manner, as exemplified by GPT \cite{radford2018improving}. Below, we review two core predictive SSL techniques for ES, \emph{masked modeling} and \emph{autoregressive modeling}, and then discuss \emph{temporal point processes}, a time-centric generative modeling framework whose predictive character and practical importance warrant inclusion alongside these SSL paradigms.

\subsubsection{Masked Modeling}
\label{subsubsec:masked-modeling}

In {\it masked} (or {\it denoising}) modeling, the learner randomly masks out certain events, event attributes, or timestamps within a sequence and aims to predict the masked portions using the remaining context. This approach is inspired by the masked language modeling (MLM) paradigm~\cite{devlin2018bert}, where tokens are hidden and the model must recover them.
For ES, this can mean \emph{masking entire events} or \emph{selectively masking} specific features (e.g., diagnosis codes in healthcare, item identifiers in e-commerce). Examples of masked SSL for ES from different domains include:
\begin{itemize}
\item Healthcare: BEHRT~\cite{li2020behrt}, BRLTM~\cite{meng2021bidirectional}, and Med-BERT~\cite{rasmy2021med} pre-train transformer models to impute masked diagnoses or procedure codes. These models mainly vary in their event embedding strategies and the information extracted from patient records, typically discretizing continuous features and summing their embeddings to form final representations.

\item E-commerce: BERT4Rec~\cite{sun2019bert4rec} applies a BERT-like approach to sequences of user-item interactions, masking items that the user actually engaged with and learning to reconstruct them.

\item Finance: BERT4Eth~\cite{hu2023bert4eth} adapts the BERT framework to transaction data from the Ethereum blockchain, using masked crypto addresses as input. Adjustments like increased masking and dropout rates address the specific properties of blockchain data.

\item Gaming: \citet{pu2022unsupervised} utilize a BERT-based framework for player representation learning in massively multiplayer online games, incorporating a confidence-guided masking strategy to focus on significant actions. Player2Vec~\cite{wang2024player2vec} goes one step further towards an LLM-based paradigm by transforming raw mobile game event logs into a customized textual vocabulary and pre-training a Longformer~\cite{beltagy2020longformer} language model with a masked objective. This setup treats player–game interaction sequences as if players were ``talking'' to the game in a domain-specific language, and yields long-context player embeddings that are useful for downstream segmentation and personalization.
\end{itemize}

Notably, further refinements to these models have emerged across domains. For example, SARD~\cite{kodialam2021deep} enhances EHR modeling by adding a reverse distillation signal during pre-training and expanding the event vocabulary. DuETT~\cite{labach2023duett} introduces a dual-attention architecture pre-trained with a masked objective. For SRS, S3Rec~\cite{zhou2020s3} extends masked modeling by introducing additional pre-training objectives around the masking paradigm, coupled with mutual information maximization as a loss function.

By focusing on partial observations and contextual relationships across timestamps, masked modeling can effectively handle the structured, high-dimensional attributes of ES data. In particular, its ability to attend to flexible contexts, rather than relying on uniform sampling intervals, makes it well-suited to irregular event arrivals that frequently occur in real-world ES scenarios.

\subsubsection{Autoregressive Modeling}
\label{subsubsec:ar-modeling}
Autoregressive (AR) approaches for ES revolve around predicting the {\it next event} (or next attribute) given the historical context. These methods extend concepts from language modeling \cite{radford2018improving}, forcing a strictly causal order -- using only past events to predict future ones.

Many SRS adopt this AR perspective, where the AR objective is commonly applied to sequences of user-item interactions, aiming to predict the next item~\cite{tang2018personalized,luo2023mcm,yuan2019simple}. While many methods focus on architectural improvements, the learning objective often remains consistent. For example, \citet{li2018learning} introduced dual recurrent networks to capture short- and long-term dependencies, while attention-based approaches~\cite{wang2020jointly,ying2018sequential} have also been employed for similar purposes.

To avoid computing normalizing constants in loss functions, techniques like noise contrastive estimation (NCE)~\cite{gutmann2010noise} have been adopted, as seen in \citet{wang2020jointly,yuan2019simple,hou2022towards}, or by measuring the distance between predicted and target item embeddings~\cite{li2018learning}. Contrastive learning approaches for SRS often incorporate AR objectives as auxiliary signals due to their alignment with the primary task~\cite{xie2022contrastive,wang2023sequential,ma2020disentangled}.

Beyond SRS, AR objectives have also been applied in gaming. \citet{zhao2021player} used AR models to generate realistic action sequences for role-playing games, leveraging player behavior patterns. \citet{kantharaju2020discovering} used AR objectives to encode gameplay sessions, enabling unsupervised strategy label discovery through embeddings. Similarly, \citet{min2016player} pre-trained action representations for open-world games using AR objectives, enhancing performance on goal recognition tasks.

Autoregressive models are intuitive for ES since many tasks, such as predicting future purchases, patient visits, or player actions, naturally align with next-event prediction. However, purely AR training can overemphasize short-horizon patterns at the expense of longer-term dependencies. To address timing gaps more faithfully, \citet{li2020time} introduced elapsed-time embeddings, providing a more nuanced temporal perspective. These efforts underscore the benefits of AR models for ES, while illustrating the need for time-aware or hybrid solutions in more complex real-world scenarios.

\subsubsection{Temporal Point Processes}
\label{subsubsec:tpp}

Temporal Point Processes (TPPs) occupy a distinct position in this taxonomy. Unlike masked and autoregressive objectives, which originated as self-supervised pretext tasks for representation learning, TPPs are a generative modeling framework designed to specify the probability distribution over when events occur. Rather than predicting the next token in a sequence, TPPs specify a conditional intensity function that depends on the history of previous events to determine the timing of future ones. They are particularly suited to domains emphasizing the temporal dimension, such as high-frequency trades in finance \cite{guo2018initiator} or retweet cascades in social media \cite{zhao2015seismic}.

Their inclusion under predictive SSL reflects both practical importance and conceptual overlap: like autoregressive methods, TPPs predict future events from history, and like SSL broadly, they learn from unlabeled data by using the event times themselves as the supervisory signal. However, an important distinction applies. In masked or autoregressive SSL, the pretext task is scaffolding for learning transferable representations, whereas in TPP training the likelihood objective is typically both the means and the end. We nonetheless include TPPs because (1) they address the same ES modeling challenges, (2) recent work explicitly integrates them with contrastive SSL objectives, (3) neural TPP encoders are increasingly reused as pre-trained components in larger pipelines, and (4) they offer the most principled probabilistic treatment of irregular event timing among the paradigms surveyed here.

Concretely, each event is represented by a timestamp $t_{u,i}$ (Equation~\ref{eq:es-def}), and in the case of marked TPPs (MTPPs), also by a discrete label $m_i$. This representation can be limiting for ES with richer metadata $\mathcal{D}$, e.g., numerical or textual features, since TPPs often restrict such attributes to a single discrete mark set $\mathcal{M}$. However, for problems where timing is paramount and events can be categorized with discrete labels, TPPs offer a principled approach to capture temporal patterns within ES data.

The pioneering work of \citet{du2016recurrent} introduced neural architectures for MTPPs, encoding event histories with recurrent neural networks and conditioning the intensity function on these representations; later, several methods have been developed for MTPPs~\cite{omi2019fully,shchur2019intensity,guo2018initiator,mei2020noise,chen2020neural,zhang2024neural}. Building on this foundation, \citet{omi2019fully} proposed a fully neural TPP model, relaxing the parametric assumptions of earlier methods, while \citet{shchur2019intensity} moved away from intensity function modeling altogether by directly estimating the distribution of inter-event times. Other approaches incorporate Noise Contrastive Estimation (NCE)~\cite{gutmann2010noise} to optimize TPP parameters without computing normalizing constants~\cite{guo2018initiator,mei2020noise}. Further innovations include:
\begin{itemize}
\item Extensions to spatial settings: Spatiotemporal point processes~\cite{chen2020neural} incorporate event locations, with $\mathcal{D} = \mathbb{R}^d$, but still focus on timing plus a continuous spatial component rather than general high-dimensional metadata.

\item Self-exciting dynamics: Neural adaptations of Hawkes processes~\cite{hawkes1971spectra}, such as the Neural Hawkes Process~\cite{mei2017neural}, capture {\it self-exciting} event behavior, where occurrences of events raise the likelihood of subsequent arrivals for a period.

\item Attention and ODE-based TPPs: Transformer adaptations~\cite{zuo2020transformer-hawkes,zhang2020self-attentive,yang2022transformer} improve modeling of long-term dependencies, while neural ODE frameworks~\cite{chen2020neural,zhang2024neural} support continuous-time hidden states for greater flexibility.

\item State-space TPPs: S2P2~\cite{chang2025deep} adapts deep state-space models (SSMs) to marked temporal point processes, interleaving stochastic jump differential equations with nonlinearities to capture continuous-time dynamics, suggesting that SSM-style inductive biases may be particularly well-suited to irregular event arrivals.

\item Combining contrastive signals: HYPRO~\cite{xue2022hypro} and the work of \citet{wang2023hierarchical} integrate {\it contrastive learning} at the sequence level to refine long-horizon TPP forecasts (see Section~\ref{subsubsec:instance-contrastive} for details).
\end{itemize}

Collectively, these TPP variants remain largely domain-agnostic and have been evaluated on diverse benchmarks, including financial transactions and social streams. A key strength is their native handling of irregular arrival patterns, which are ubiquitous in real-world ES: unlike methods that assume regularly sampled data, TPPs model continuous time directly. However, as noted above, the restriction to discrete marks limits their ability to capture richer, multi-dimensional event attributes. Consequently, if the ultimate goal is task-agnostic representation learning (for instance, for clustering patients or analyzing user behavior broadly), TPPs may need complementary mechanisms or hybrid approaches to fully leverage metadata beyond event timing and discrete categories.


\subsection{Contrastive SSL}
\label{subsec:contrastive-ssl}

Contrastive SSL methods learn a representation space where \emph{similar} event streams or subsequences are pulled together, while \emph{dissimilar} ones are pushed apart \cite{liu2021self}. This high-level approach typically avoids reconstructing low-level details, which may be noisy or irrelevant, and can thereby learn more robust \emph{entity-level} embeddings. As illustrated in Figure~\ref{fig:ssles-tree-diagram}, contrastive methods remain relatively underutilized in ES, yet they offer strong potential across diverse application settings.

\subsubsection{Instance Contrastive}
\label{subsubsec:instance-contrastive}

Instance-based contrastive learning aims to learn discriminative representations by treating different augmentations of the \emph{same} event stream instance as \emph{positive} pairs and instances (or their augmentations) from \emph{other} entities as \emph{negative} pairs. This paradigm can be traced back to frameworks like SimCLR~\cite{chen2020simple} and MoCo~\cite{he2020momentum}, which rely on semantic-preserving data transformations, or augmentations, to drive representation learning. Formally, if $\mathcal{A}$ is a set of possible augmentations, we draw two random transformations $a_1, a_2 \sim \mathcal{A}$ and apply them to an event stream $S_u$. The resulting views $a_1(S_u) = {S}^{(1)}_u$ and $a_2(S_u) = S^{(2)}_u$ form a {\it positive} pair, while event streams from other entities play the role of {\it negative} samples.

Unlike computer vision, where image augmentations include simple pixel-level transformations such as rotation or color jitter, ES data is {\it irregular} and {\it discrete}, often containing complex contextual attributes.

Existing augmentation strategies for ES can be broadly grouped into three categories: (1) temporal augmentations, which modify the time dimension through subsequence cropping, time shifting, or resampling at different granularities~\cite{babaev2022coles,jeong2023event,ma2020disentangled}; (2) structural augmentations, which alter event composition through random event dropout, feature masking, or attribute perturbation~\cite{xie2022contrastive,hou2022towards,raghu2023sequential}; and (3) compositional augmentations, such as item reordering or sequence mixing, which recombine elements while preserving overall semantics~\cite{xie2022contrastive,wang2023sequential}. Determining which transformations preserve the underlying semantics of an ES remains an open problem: for instance, randomly dropping medical events may destroy clinically meaningful patterns, while aggressive temporal cropping in e-commerce may sever the connection between browsing and purchase behavior. This challenge has motivated interest in learned augmentation policies that adapt transformations to the data distribution, though such approaches remain largely unexplored in ES contexts.

For irregular and multimodal ES, augmentation design must additionally respect temporal coherence and cross-modal consistency. When event timing is semantically meaningful, operations such as cropping, masking, or mild time jittering should not invert causal orderings or create impossible gaps. For multimodal events (e.g., vitals, lab tests, and clinical notes at the same time point), transformations need to be applied in a time-locked fashion across modalities so that different views of the same event remain coherent. Finally, augmentations should avoid both label leakage and unrealistic artifacts, especially in domains where downstream tasks depend on absolute times or rare event types.
Several representative instance contrastive methods have emerged across different ES domains:
\begin{itemize}
\item Sequential recommenders: CL4SRec~\cite{xie2022contrastive} combines event masking, subsequence cropping, and item reordering to generate positive pairs of user interaction sequences, while negative samples come from other users. Other works~\cite{qiu2022contrastive,hou2022towards,wang2023sequential} adapt similar ideas, sometimes adding \emph{supervised} pair-mining (e.g., user sessions ending in the same item).

\item Finance: CoLES~\cite{babaev2022coles} proposes random subsequence sampling for transaction logs, using the InfoNCE objective to pull together augmentations of the same account’s activity while pushing apart different accounts.

\item Healthcare: \citet{raghu2023sequential} mask entire modalities (e.g., vitals vs.~labs) and add noise to EHR data, employing the SimCLR-based NT-Xent loss for contrast. \citet{jeong2023event} define positive pairs as prefix and suffix segments around critical hospital visits.

\item Combining TPP: \citet{wang2023hierarchical} and \citet{xue2022hypro} combine TPP's autoregressive objectives with an instance contrastive signal to differentiate authentic event continuations from artificially generated ones.
\end{itemize}

Instance-based contrastive SSL is conceptually straightforward and avoids the need to reconstruct potentially noisy event attributes, focusing instead on alignment of semantically consistent representations. This can yield robust entity-level embeddings for complex, high-dimensional streams. However, several practical challenges persist:
\begin{itemize}
\item Defining valid augmentations: ES data is irregular and domain-specific, making it non-trivial to design transformations (e.g., reordering vs. partial masking) that preserve critical chronological or contextual relationships.
\item Negative sampling: Methods must ensure negative pairs genuinely represent different entities or distinct behaviors, especially if multiple entities have partially similar sequences.
\item Data sparsity: Sparse or short event sequences limit opportunities for subsequence-based augmentations or heavier masking without losing crucial information.
\end{itemize}
Despite these hurdles, instance-based contrastive learning remains a promising strategy for leveraging unlabeled ES in diverse domains, with growing evidence that carefully chosen augmentations and objectives can substantially improve downstream performance.

\subsubsection{Distillation}
\label{subsubsec:distillation-methods}

Distillation-based contrastive learning avoids explicit negative samples by coupling a {\it teacher} network and a {\it student} network on differently augmented views of the same event stream. Inspired by methods such as BYOL~\cite{grill2020bootstrap}, SimSiam~\cite{chen2021exploring}, and DINO~\cite{caron2021emerging}, the student aims to predict the teacher’s latent representations. The construction of the target branch varies across these frameworks. SimSiam shares encoder parameters between the two branches and stops gradients through the target representation~\cite{chen2021exploring}, whereas BYOL and DINO update the teacher parameters using an exponential moving average of the student's weights~\cite{grill2020bootstrap,caron2021emerging}. These mechanisms help avoid collapse (i.e., trivial solutions) and enable the model to learn robust embeddings without requiring negative pairs.

Distillation-based approaches remain relatively rare in ES contexts. One notable example is Hi-BEHRT~\cite{li2022hi-behrt}, which builds on the BEHRT framework for EHR data~\cite{li2020behrt}. In Hi-BEHRT, two augmented views of a patient's event sequence are created (e.g., via event masking or subsequence sampling). The teacher and student networks independently encode these augmented sequences, and a BYOL-style distillation loss forces the student’s latent representation to match that of the teacher. By circumventing the need for negative sampling, Hi-BEHRT can focus more directly on capturing meaningful patterns in clinical data, particularly when event streams are irregular or domain-specific.

Distillation-based frameworks avoid explicit negative sampling, making them appealing in domains where negative examples are ambiguous or tedious to define. Nevertheless, designing effective augmentations (e.g., event masking or subsequence cropping) remains a key challenge, and the teacher-student alignment can lead to overfitting if spurious correlations are learned.


\subsubsection{Feature Decorrelation}
\label{subsubsec:feature-decorrelation}

Feature decorrelation methods aim to learn representations with {\it minimal redundancy} across latent dimensions, typically without explicit negative pairs. Frameworks like Barlow Twins~\cite{zbontar2021barlow} and VICReg~\cite{bardes2021vicreg} proceed by generating two augmented views of the same sequence and passing them through a shared model. Although both methods align augmented views while discouraging redundant features, their loss formulations differ. Barlow Twins uses the diagonal and off-diagonal terms of the cross-correlation matrix to promote alignment and redundancy reduction, whereas VICReg combines three explicit terms for invariance, variance, and covariance.

Feature decorrelation remains much less explored than predictive or instance-based contrastive methods in ES contexts. One notable exception is \citet{raghu2023sequential}, who adopt a VICReg-type loss for healthcare EHR data. Their method processes two augmented views (e.g., partial masking or noise injection across different clinical modalities) and learns embeddings via alignment plus decorrelation. Although originally demonstrated in a multimodal EHR setting, it can be adapted to purely structured or partially unstructured event streams, thereby highlighting its potential generality.

Similar to distillation, feature decorrelation eliminates the need for negative samples. The two approaches differ in mechanism: decorrelation operates on encoder output statistics (e.g., cross-correlation matrices), whereas distillation aligns student predictions to teacher representations. They are complementary and could be combined, though this direction remains unexplored in ES contexts. The success of decorrelation still relies on carefully designed augmentations and balancing alignment with redundancy reduction.


\subsubsection{Multimodal Contrastive}
\label{subsubsec:multimodal-contrastive}

Multimodal contrastive learning aims to align representations across multiple modalities associated with the same data instance, ensuring that semantically related information from different sources is embedded closer together in the learned space. This approach is commonly exemplified by models like CLIP~\cite{radford2021learning}, which aligns image-text pairs by training encoders for each modality with a shared contrastive loss. Typically, a positive pair consists of aligned representations (e.g., image and caption), while unrelated data serves as negative pairs. This setup can be adapted for ES data when events are enriched with multimodal features.


Event streams in domains like healthcare often contain inherently multimodal data, such as time-series vital signs, diagnostic codes, and accompanying textual notes or summaries. These rich contexts make multimodal contrastive learning a natural fit, though its application remains limited due to the preprocessing required to align modalities effectively.
For instance, \citet{king2023multimodal} explored a multimodal contrastive objective inspired by CLIP for Intensive Care Unit (ICU) data, where time-series signals from medical monitors were contrasted with clinical notes. This event-level alignment enabled the model to integrate structured numeric data with unstructured text, providing a more holistic representation of patient states. Extending this idea, \citet{ma2024global} applied multimodal contrastive learning to align complete ICU stays. Their approach leveraged discharge summaries as high-level textual representations of the patient’s trajectory, contrasting these with aggregated time-series features from the entire hospital stay. While this extension captured broader temporal contexts, it also required comprehensive summaries as a secondary modality, which may not always be available.

Multimodal contrastive methods offer the significant advantage of integrating diverse data types, enabling models to capture richer and more holistic event representations. However, their success relies on the availability and alignment of multiple modalities, which can introduce challenges in preprocessing and data curation. Additionally, the reliance on contrastive objectives across modalities may lead to representational biases if one modality dominates. Nevertheless, for ES domains with naturally multimodal data, such as healthcare, these methods show considerable promise in bridging gaps between structured and unstructured event attributes. Algorithmically, multimodal contrastive learning is a cross-modal specialization of instance-based contrast, distinguished by view heterogeneity rather than a fundamentally different objective.


\subsection{Summary}
\label{subsec:ssl-summary}

The reviewed literature underscores that \textit{predictive methods} dominate SSL for ES due to their natural alignment with the sequential and contextual structure of event streams. Masked modeling, inspired by advancements in NLP, excels at reconstructing missing event attributes and is widely applied across domains such as healthcare, e-commerce, gaming, and finance. Autoregressive approaches similarly align pre-training goals with downstream tasks, making them particularly effective in sequential recommendation systems and time-sensitive applications. TPPs, though originating as generative models rather than representation learning techniques, complement these objectives as a time-centric framework: they offer mathematically rigorous tools for modeling irregular event timing, and recent work increasingly integrates them with contrastive SSL objectives or reuses their encoders in broader pipelines.

Contrastive methods remain relatively underexplored, despite their potential to produce robust, entity-level representations. Instance-based contrastive learning has been applied to healthcare, finance, and recommender systems, leveraging augmentations such as subsequence sampling and event masking. Distillation-based contrastive methods avoid the complexity of negative sampling but require well-designed teacher-student alignments and effective augmentations. Feature decorrelation methods, while promising for high-dimensional ES data, have seen limited applications and demand further exploration. Emerging multimodal contrastive techniques, particularly in healthcare, show great promise in integrating diverse modalities such as time-series vitals and textual notes, though their success hinges on data availability and preprocessing.

As seen in Figure~\ref{fig:ssles-tree-diagram}, predictive approaches are dominant across domains, while contrastive and hybrid paradigms remain nascent but hold significant potential. Advancing SSL for ES modeling requires addressing key gaps, including the need for unified frameworks, scalable methodologies, and better benchmarking resources. We believe future work should prioritize exploring underutilized paradigms, such as multimodal and decorrelation-based methods, and integrating insights from both predictive and contrastive approaches to create robust, domain-agnostic SSL solutions for event streams. Combining paradigms, for instance, contrastive and masked objectives within a single model, remains largely unexplored and represents a promising research direction.

Given the current state of the literature, we offer guidance for practitioners selecting among SSL paradigms for ES applications. This guidance reflects observed patterns rather than systematic empirical comparisons, which remain limited.

When the downstream task involves next-event prediction, autoregressive objectives offer natural alignment, as the pre-training and evaluation objectives are structurally similar; this is the dominant choice in sequential recommendation. When the task requires holistic sequence understanding, such as classification, risk prediction, or outcome regression, masked modeling provides bidirectional context that captures dependencies across the full sequence; this approach prevails in healthcare applications. When the goal is to produce entity-level representations for clustering, retrieval, or similarity-based applications, contrastive methods avoid reconstruction of potentially noisy low-level details and focus on discriminative embeddings; however, they require careful augmentation design. When event timing is the primary quantity of interest, e.g., for predicting inter-arrival gaps or modeling self-exciting dynamics, temporal point processes offer a principled probabilistic framework, though at the cost of flexibility in handling rich event metadata.

A further practical consideration is the cold-start setting, where new entities arrive with short histories and events fall outside the training vocabulary, such as newly introduced items, event types added by game updates, or newly adopted medical codes. Masked and autoregressive models with fixed event vocabularies must be re-tokenized or extended when new event types appear, and TPPs with a fixed mark set cannot score unseen marks. Contrastive methods sidestep the vocabulary issue at the sequence level, but degrade for entities whose histories are too short to augment (Section~\ref{subsubsec:instance-contrastive}). Approaches that compose events from textual descriptions mitigate out-of-vocabulary events by construction, at the cost of longer input sequences (Section~\ref{subsec:llm-fm-es}). Cold-start robustness is rarely evaluated explicitly in the surveyed literature and remains an open problem.

Beyond task alignment, data and domain characteristics also inform paradigm choice. For short sequences with limited context, masked modeling may remove too much signal, favoring autoregressive or contrastive approaches. For noisy or sparse event attributes, such as in gaming, where players may perform repetitive actions, contrastive methods that operate at the entity level may be more robust than reconstruction-based objectives. For multimodal events combining structured and unstructured data, as frequently encountered in healthcare, multimodal contrastive alignment (Section~\ref{subsubsec:multimodal-contrastive}) can integrate heterogeneous information, though preprocessing requirements increase. These considerations are summarized in Table~\ref{tab:paradigm-guidance}.

We emphasize that these recommendations are heuristic: the field lacks controlled experiments comparing SSL paradigms across tasks and domains under matched conditions. Developing such systematic comparisons remains an important direction for future work. A minimal such comparison would hold the encoder architecture, pre-training corpus, and compute budget fixed while varying only the pretext objective. The resulting representations would then be evaluated on a shared battery of downstream tasks spanning at least two domains, under matched fine-tuning and labeling regimes. The standardized protocols of EBES~\cite{osin2025ebes} offer a natural starting point.

\begin{table}[ht!]
    \caption{Qualitative guidance for SSL paradigm selection in ES applications, based on currently observed patterns in the literature.}
    \label{tab:paradigm-guidance}
    \centering
    \footnotesize
    \renewcommand{\arraystretch}{1.4}
    \setlength{\tabcolsep}{4pt}
    \begin{tabular}{>{\raggedright\arraybackslash}p{0.22\linewidth}|
                    >{\raggedright\arraybackslash}p{0.35\linewidth}|
                    >{\raggedright\arraybackslash}p{0.35\linewidth}}
        \textbf{Paradigm} & \textbf{Well-suited scenarios} & \textbf{Considerations} \\ \hline
        Autoregressive & Next-event/next-item prediction; recommendation; trajectory forecasting & May overemphasize short-horizon patterns; natural task alignment simplifies fine-tuning \\ \hline
        Masked modeling & Classification; regression; risk prediction; tasks requiring bidirectional context & Masking ratio and strategy require tuning; may struggle with very short sequences \\ \hline
        Temporal point processes & Inter-arrival time prediction; self-exciting dynamics; timing-centric tasks & Limited flexibility for rich metadata; strong probabilistic grounding \\ \hline
        Instance contrastive & Entity-level clustering; retrieval; similarity-based tasks; noisy event attributes & Augmentation design is critical; avoids low-level reconstruction \\ \hline
        Multimodal contrastive & Events with heterogeneous modalities (e.g., vitals + clinical notes) & Requires aligned multimodal data; preprocessing overhead
    \end{tabular}
\end{table}

\section{Datasets and Downstream Tasks}
\label{sec:es-data-overview}
The success of SSL for ES modeling depends on access to robust datasets and well-defined downstream tasks. Following the standard SSL paradigm, ES methods typically proceed in two phases: (1) a pre-training phase, where models learn representations from unlabeled event sequences using pretext tasks, and (2) an evaluation phase, where the learned representations are assessed on labeled downstream tasks. Evaluation commonly takes the form of fine-tuning on task-specific data, though few-shot and zero-shot protocols are also employed. Reporting practices vary considerably across the literature, complicating cross-study comparisons. This section provides an overview of widely used public datasets (Section~\ref{subsec:datasets-overview}) and discusses downstream tasks along with current evaluation practices (Section~\ref{subsec:downstream-tsks}).

\begin{table}[h!]
\caption{Overview of commonly used benchmark datasets in ES modeling. Entity counts are approximate and may vary by version or preprocessing. For StackOverflow and Retweet, no specific domain is assigned, as they are treated primarily as generic event sequences.}
\label{tab:benchmark-datasets}
\centering
\scriptsize
\renewcommand{\arraystretch}{1.3}
\begin{tabular}{>{\centering\arraybackslash}p{0.13\linewidth}|
                >{\centering\arraybackslash}p{0.10\linewidth}|
                >{\centering\arraybackslash}p{0.15\linewidth}|
                >{\centering\arraybackslash}p{0.22\linewidth}|
                >{\arraybackslash}p{0.30\linewidth}}
\textbf{Name} & \textbf{Domain} & \textbf{Entities} & \textbf{Used in} & \textbf{Description} \\ \hline
\textcolor{cbGreen}{CPRD} \cite{herrett2015data-cprd} & Healthcare & 60M patients & \citet{li2020behrt,li2022hi-behrt} & Medical events, tests, and diagnoses from patients' primary care visits. \\ \hline
\textcolor{cbGreen}{MIMIC} \cite{saeed2002mimic,johnson2016mimic,johnson2023mimic-iv} & Healthcare & 365K patients & \citet{jeong2023event,raghu2023sequential,king2023multimodal,labach2023duett,enguehard2020neural,mei2017neural,zhang2020self-attentive,yang2022transformer,du2016recurrent} & Medical tests and vitals from patients, primarily in ICU settings. \\ \hline
\textcolor{cbBlue}{Amazon} \cite{mcauley2015image} & \mbox{E-commerce} & 20M users & \citet{xie2022contrastive,sun2019bert4rec,zhou2020s3,ma2020disentangled,qiu2022contrastive,li2020time,wang2023sequential} & Product reviews from a wide range of categories. \\ \hline
\textcolor{cbBlue}{Yelp}\tablefootnote{\url{https://www.yelp.com/dataset}} & \mbox{E-commerce} & 2M users & \citet{xie2022contrastive,zhou2020s3,qiu2022contrastive,wang2023sequential} & Business reviews submitted by users. \\ \hline
\textcolor{cbBlue}{MovieLens} \cite{harper2015movielens} & \mbox{E-commerce} & 6K--162K users & \citet{sun2019bert4rec,ma2020disentangled,tang2018personalized,qiu2022contrastive,li2020time} & Users rating and tagging movies. \\ \hline
StackOverflow \cite{snapnets} & -- & 6K sequences & \citet{mei2017neural,zhang2020self-attentive,yang2022transformer,zuo2020transformer-hawkes,wang2023hierarchical,xue2022hypro,du2016recurrent} & User reward history treated as a sequence, with events signifying individual rewards. \\ \hline
Retweet \cite{zhao2015seismic} & -- & 24K sequences & \citet{zuo2020transformer-hawkes,zhang2020self-attentive,wang2023hierarchical,guo2018initiator} & Retweet cascades from original tweets on Twitter. \\
\end{tabular}
\end{table}

\subsection{Overview of Public Datasets}
\label{subsec:datasets-overview}
ES datasets are categorized by their source applications, such as healthcare, e-commerce, and generic event streams. Table~\ref{tab:benchmark-datasets} lists commonly used datasets, their domains, and descriptions.

In healthcare, prominent datasets include CPRD \cite{herrett2015data-cprd} and MIMIC \cite{saeed2002mimic,johnson2016mimic,johnson2023mimic-iv}. CPRD provides primary care records from UK general practice visits, while the series of MIMIC datasets offer detailed ICU patient data, including medical tests and vital signs. These datasets have supported methods such as BEHRT~\cite{li2020behrt}, Hi-BEHRT~\cite{li2022hi-behrt}, and multimodal contrastive learning~\cite{king2023multimodal}. However, preprocessing these datasets often requires converting event streams into time series or tabular formats, which can result in information loss, especially when binning irregular event times~\cite{wang2020mimic-extract,mcdermott2021comprehensive}.

In e-commerce, datasets like Amazon~\cite{mcauley2015image}, Yelp, and MovieLens~\cite{harper2015movielens} are frequently used in SRS. These datasets typically focus on user reviews or ratings rather than clicks or purchases, leading to a divergence between the training data and real-world interaction tasks. Methods such as BERT4Rec~\cite{sun2019bert4rec}, CL4SRec~\cite{xie2022contrastive}, and ContraRec~\cite{wang2023sequential} have adapted these datasets to SSL paradigms.

For generic event streams, StackOverflow~\cite{snapnets} and Retweet~\cite{zhao2015seismic} are widely used. StackOverflow captures user reward histories, while Retweet records cascades of social media interactions. These datasets serve as benchmarks for TPPs~\cite{mei2017neural,zhang2020self-attentive,xue2022hypro}, focusing on the irregular timing of events.

Despite the availability of these datasets, challenges remain. Public data availability varies considerably across domains: healthcare datasets represent carefully de-identified releases from academic institutions, while e-commerce datasets primarily consist of reviews and ratings rather than granular behavioral logs. Gaming and finance are notably underrepresented, as event logs in these domains are typically business-sensitive and subject to regulatory constraints, forcing research to rely on proprietary data, which are inaccessible to the broader research community~\cite{min2016player,min2017multimodal,wang2024player2vec}. Moreover, the lack of standardized preprocessing pipelines and benchmarks, as seen in domains like healthcare~\cite{gupta2022extensive}, hampers reproducibility and model comparison.

\subsection{Downstream Tasks}
\label{subsec:downstream-tsks}
SSL methods for ES aim to pre-train models that can be fine-tuned or directly applied to real-world tasks. Downstream tasks for ES can be broadly categorized into several types: (1) classification tasks, such as mortality prediction, fraud detection, or churn prediction, which assign discrete labels to entities or events; (2) regression tasks, such as length-of-stay estimation or risk scoring, which predict continuous outcomes; (3) ranking and retrieval tasks, such as next-item recommendation, which order candidate items by relevance to a query or context; and (4) generation tasks, such as future event prediction or trajectory forecasting, which produce sequences of subsequent events.
The alignment between pre-training objectives and downstream tasks plays a notable role in transfer performance. Autoregressive pre-training naturally aligns with next-event prediction and recommendation, as both involve forecasting future elements from historical context. Masked modeling, by capturing bidirectional context, tends to benefit classification and regression tasks that require holistic sequence understanding. Contrastive methods, which yield discriminative entity-level representations, are particularly suited for clustering, retrieval, and similarity-based applications. Despite these intuitions, systematic empirical studies of pretext-task alignment remain limited, and most published works evaluate on a narrow set of tasks within a single domain. The following paragraphs summarize common downstream tasks and evaluation practices across the domains covered in this survey.

In healthcare, tasks often involve predicting patient outcomes, such as risk stratification, disease diagnosis, or mortality prediction. For example, BEHRT~\cite{li2020behrt} and Hi-BEHRT~\cite{li2022hi-behrt} have been evaluated on tasks like predicting hospitalizations or the onset of chronic conditions. Metrics include precision, recall, and area under the receiver operating characteristic (AUROC) curve. Pre-training typically involves masked modeling or contrastive SSL to generate robust patient embeddings that can generalize across various prediction tasks~\cite{rasmy2021med,raghu2023sequential}.

In e-commerce, downstream tasks align closely with SRS objectives, such as next-item prediction or user behavior forecasting. Pre-trained models like BERT4Rec~\cite{sun2019bert4rec} and ContraRec~\cite{wang2023sequential} fine-tune on datasets like Amazon or Yelp to optimize metrics such as hit rate (HR), normalized discounted cumulative gain (NDCG), and mean reciprocal rank (MRR). These tasks often rely on autoregressive or instance-contrastive objectives.

Gaming tasks focus on player behavior modeling, such as goal prediction or style classification. While gaming datasets are typically proprietary, models like Player2Vec~\cite{wang2024player2vec} demonstrate how LLM-style SSL pre-training on tokenized event sequences can infer meaningful player representations for tasks like churn prediction or personalized matchmaking.

In finance, tasks such as fraud detection or transaction de-anonymization leverage pre-trained blockchain transaction representations. BERT4Eth~\cite{hu2023bert4eth} fine-tunes masked modeling outputs on phishing detection datasets, evaluated using precision-recall trade-offs.

Across these domains, SSL pre-training strives to align pretext tasks with real-world objectives, enabling robust transfer learning. However, several factors limit the conclusions that can be drawn from current evaluation practices. First, most works evaluate within a single domain and on a narrow set of tasks, making it difficult to disentangle improvements due to the SSL method from those attributable to domain-specific design choices. Second, evaluation protocols vary widely: some studies report fully fine-tuned performance, others assess few-shot or zero-shot capabilities, and the amount of labeled data used during fine-tuning is often inconsistent or underspecified. Third, the metrics employed differ across domains (e.g., AUROC in healthcare, hit rate in recommendation), complicating cross-domain comparisons even when methods are conceptually similar. These limitations underscore the need for standardized evaluation frameworks that assess SSL methods across multiple tasks, domains, and labeling regimes, as discussed further in Section~\ref{subsec:open-datasets}.

\section{Progress and Prospects}
\label{sec:progress-prospects}

The rapid progress and evolution of SSL for ES modeling have highlighted both promising directions and key gaps. In the following subsections, we (1) explore the potential for domain-agnostic SSL frameworks for ES modeling (Section~\ref{subsec:domain-agnostic-learning}), leveraging structural similarities across diverse application domains, (2) discuss underexplored modeling paradigms (Section~\ref{subsec:underexplored-modern-paradigms}), (3) outline the critical aspect of timestamp information (Section~\ref{subsec:event-time-modeling}), (4) highlight the need for open-source benchmarks to enable progress in the field (Section~\ref{subsec:open-datasets}) and (5) assess the current state and limitations of LLM-style and foundation models for ES (Section~\ref{subsec:llm-fm-es}).

\subsection{Domain-Agnostic Learning}
\label{subsec:domain-agnostic-learning}
The review of SSL methods for ES modeling highlights their broad applicability across diverse domains. Despite this diversity, many approaches converge in terms of architecture and learning objectives, which can be attributed to structural similarities inherent in the data across these fields. Domains such as recommender systems, game event modeling, and EHRs involve sequential event data generated by discrete entities, including users, players, or patients. Individual interactions, such as user-item engagements, in-game actions, or medical assessments, are treated as discrete events, emphasizing the potential for developing generalized SSL methods for ES data.


The consistent structure of ES data and the alignment of learning objectives across domains present an opportunity to design generalized SSL frameworks for ES modeling. Such domain-agnostic approaches could pave the way for foundational models in ES, where further research can prioritize methods that are adaptable and generalizable across a range of applications. This direction holds promise for creating unified modeling strategies that not only bridge gaps between specific domains but also leverage shared patterns within ES data for more robust, scalable ES modeling methodologies based on SSL. These considerations are closely related to the emerging discussion around large language and foundation models for temporal data; we return to this point in Section~\ref{subsec:llm-fm-es}, where we discuss LLM-style approaches and the current limits of foundation models for ES.

\subsection{Underexplored Modern Paradigms}
\label{subsec:underexplored-modern-paradigms}
Research in SSL has made significant progress across various domains, yet the majority of methods in ES remain focused on predictive learning objectives, often inspired by NLP. This narrow focus leaves other promising directions, such as contrastive learning, relatively unexplored, despite its demonstrated success in other areas~\cite{chen2020simple,radford2021learning,niizumi2021byol,oquab2023dinov2}.

Contrastive learning operates on high-level concepts and avoids reliance on reconstructing low-level details, such as pixels or tokens, during training. This reduces sensitivity to noise in low-level features~\cite{liu2021self}, a desirable trait for producing high-quality representations~\cite{bengio2013representation}. In ES data, these high-level objects are often entities, making entity-level modeling particularly valuable, as downstream tasks typically require robust representations of entities like patients in healthcare or players in gaming, rather than individual events. Individual events, with their diverse metadata and potential noise (e.g., misclicks on items), can introduce a noisy signal during pre-training, further highlighting the advantages of focusing on higher-level representations.
However, contrastive SSL paradigms remain underutilized in ES modeling. This gap may stem from factors such as the complexity of implementing contrastive methods~\cite{balestriero2023cookbook}, challenges in defining semantic-preserving augmentations for ES, or the similarity of ES data to language data, where predictive objectives are more commonly used~\cite{liu2021self}. By contrast, predictive objectives can often be adopted almost verbatim from NLP: tokenization schemes, architectures, training recipes, and even off-the-shelf implementations transfer directly to event sequences, lowering the engineering barrier and aligning neatly with common downstream tasks. This asymmetry helps explain the current research landscape and highlights the need for more systematic work on contrastive designs that are tailored to event data. A shift toward contrastive learning could enable the development of models that capture stable, generalizable representations of data-generating entities, aligning more effectively with downstream task requirements.

Novel approaches like the Joint Embedding Predictive Architecture (JEPA)~\cite{lecun2022path} also present promising opportunities, as demonstrated in other modalities~\cite{assran2023self,fei2023a-jepa,skenderi2023graph}. By focusing on latent-space representations and predicting abstract features rather than low-level details, JEPA-based methods reduce sensitivity to noise in individual events while maintaining the integrity of learned representations~\cite{bardes2024revisiting,assran2023self,littwin2024jepa}. These latent-space approaches offer a pathway to more robust and generalizable models suited to a variety of tasks.

Future research should explore these alternative paradigms to better capture entity-level representations in ES data. Expanding beyond predictive objectives has the potential to create models that handle complex, high-dimensional event streams more effectively, ultimately enhancing performance across diverse downstream tasks.

\subsection{Event Time Modeling}
\label{subsec:event-time-modeling}
The treatment of event timestamps in event stream modeling varies significantly, with some methods disregarding timestamps entirely, as seen in sequential recommendation systems~\cite{zhou2020s3,tang2018personalized}, while others explicitly model temporal dynamics through TPPs~\cite{mei2017neural,omi2019fully}. Another approach incorporates timestamps as contextual features, particularly in EHR modeling~\cite{li2020behrt,rasmy2021med}. Despite these strategies, the role of timestamps in self-supervised ES modeling remains underexplored, with limited systematic evaluation of their influence on model performance.

In predictive SSL, the decision to mask timestamps alongside masked events introduces open questions. Masking temporal information could either reveal unintended dependencies or enhance generalizability, but the impact on tasks requiring temporal consistency remains unclear. These considerations have yet to be thoroughly investigated across methods.

Future work should systematically examine timestamp modeling to establish best practices and evaluate the contribution of temporal information to model robustness. Initial studies could focus on the effects of timestamp masking, providing insights into how temporal data should be integrated into SSL-based ES frameworks.

\subsection{Open Datasets and Benchmarks}
\label{subsec:open-datasets}
Until recently, progress in SSL for ES was hindered by the limited availability of open-source datasets and the absence of standardized benchmarks spanning multiple domains~\cite{shchur2021neural}. Most widely used datasets were concentrated in healthcare and recommendation systems, leaving other domains underrepresented. As a result, researchers often repurpose unrelated datasets like StackOverflow~\cite{snapnets} and Retweet~\cite{zhao2015seismic} for ES tasks, which leads to suboptimal benchmarking and limits the generalizability of findings. This situation has begun to improve: EBES~\cite{osin2025ebes} provides the first comprehensive benchmarking framework for event sequence classification, featuring standardized evaluation protocols, a curated collection of datasets, and unified implementations of baseline models. However, EBES focuses on sequence-level classification and regression tasks; broader SSL objectives such as representation learning, few-shot transfer, and generative modeling remain without standardized evaluation, and coverage of domains beyond healthcare, finance, and e-commerce is limited.

This issue is further compounded by the narrow range of downstream tasks associated with existing benchmarks. Evaluations often focus on a limited set of objectives, restricting the scope of ES modeling assessments and hindering the development of versatile SSL methods. Recent benchmarking efforts have begun to address the lack of systematic empirical comparisons. EBES~\cite{osin2025ebes} evaluates multiple architectures under controlled hyperparameter optimization and cross-validation protocols, revealing that model rankings vary substantially across datasets and that some widely used benchmarks may not reliably discriminate between methods. However, EBES focuses on supervised classification; most SSL-specific techniques, particularly contrastive and hybrid objectives, remain outside its scope, and cross-domain transfer evaluations are still largely absent from the literature. Extending such efforts to cover SSL paradigms would help validate the qualitative guidance offered in Section~\ref{subsec:ssl-summary} and transform current heuristic recommendations into empirically grounded best practices.

Furthermore, reliance on proprietary or private datasets~\cite{wornow2023ehrshot,zhang2018patient2vec,meng2021bidirectional,kodialam2021deep,wang2024player2vec,zhao2021player} impedes reproducibility and limits broader validation and extension of research contributions. The scarcity of datasets is likely due to the nature of the data, as it is often generated by human actions, such as healthcare patients, e-commerce customers, or game players, is often considered business-sensitive or private. Stringent regulations, including data protection laws and confidentiality agreements, further restrict the sharing of such data, creating additional challenges for developing open and diverse benchmarks.

Building on recent progress, future work should extend existing benchmarks to cover SSL-specific evaluation scenarios, while expanding domain coverage to underrepresented areas and emerging applications such as gaming and cyber-security. Expanding evaluation scopes and fostering data-sharing initiatives will enhance transparency, reproducibility, and the development of generalizable SSL approaches for ES modeling. 
At a minimum, useful synthetic or open ES benchmarks should (1) reproduce realistic temporal statistics such as burstiness and heavy-tailed inter-arrival times and event-type frequencies, (2) support multi-entity structure with heterogeneous attributes (categorical and numerical, and, where possible, additional modalities), and (3) provide a small set of well-defined downstream labels (e.g., risk, anomaly, or recommendation targets), ideally with controlled distribution shifts across domains or sites to test transfer.

Synthetic benchmarks are particularly attractive for domains where real logs cannot be released, such as DevOps operations~\cite{xie2024logsd} and cyber‑attack detection~\cite{wu2025robust}, because they can preserve realistic data distribution while avoiding disclosure of sensitive details. Recent advances in synthetic data generation techniques could enable the development of datasets that retain the properties of the original data while safeguarding confidentiality. At the same time, synthetic benchmarks carry inherent limitations: they may fail to capture rare event types or domain-specific noise patterns that are difficult to specify generatively, and they cannot fully substitute for validation against real-world outcomes. Recent toolkits have begun to address evaluation standardization for specific ES subproblems. For instance, EasyTPP~\cite{xue2024easytpp} provides a unified framework for temporal point process evaluation, standardizing data preprocessing and metrics across commonly used datasets. Extending such efforts to broader, cross-domain ES evaluation remains an important open challenge.

\subsection{Large Language and Foundation Models for Event Streams}
\label{subsec:llm-fm-es}

LLMs and so-called foundation models have transformed research in natural language processing, vision, and, more recently, structured data such as time series. In the context of ES, however, their impact is still limited. Most existing methods can be viewed as using ``LLM-style'' architectures (e.g., Transformers with masked or autoregressive SSL objectives) as building blocks for self-supervised pre-training, but without yet reaching the scale, cross-domain coverage, or emergent capabilities that are typically associated with foundation models in other modalities.

Furthermore, only a few works explicitly treat ES as a ``language'' by mapping events to discrete tokens and pre-training a sequence model on the resulting corpus. For example, Player2Vec~\cite{wang2024player2vec} transforms raw mobile game event logs into a customized textual vocabulary and trains a Longformer model on these sequences. In sequential recommendation, methods such as P5~\cite{geng2022recommendation} similarly reformulate recommendation as natural language generation, expressing user histories and predictions as text sequences processed by pretrained language models. These approaches demonstrate that language models can capture nuanced behavioral patterns and produce useful entity representations without task-specific labels. However, in healthcare, a critical review of 84 foundation models trained on non-imaging EHR data found that most are trained on small, narrowly scoped datasets and evaluated on tasks that provide limited insight into their real-world utility~\cite{wornow2023shaky}, suggesting that this paradigm remains in an early and fragmented stage across domains.

In parallel, there is a rapidly growing literature on foundation models for time series, as reviewed, for example, by \citet{liang2024foundation}. These models typically pretrain on large collections of regularly sampled sequences and aim to transfer across forecasting and classification tasks. While such work shows that foundation-style pre-training is feasible for structured temporal data, it remains an open question to what extent these models can be directly applied to more general ES. Converting ES to time series (e.g., via binning or aggregation) often entails substantial information loss and may fail to capture irregular timing or event-level metadata that are central to ES applications.

Beyond these data and evaluation constraints, LLM-style approaches for ES data face several challenges that go beyond those in standard text modeling, which may explain their limited adoption thus far. First, ES data are inherently heterogeneous: events may combine categorical codes, numerical measurements, and even unstructured content, making it nontrivial to design a single, semantically coherent tokenization scheme. Second, ES often involve multiple interacting entities (e.g., users, items, accounts), so straightforward linearization into a one-dimensional token sequence can obscure relational structure. Third, long-range dependencies in ES can easily exceed typical context windows, especially when modeling lifelong trajectories such as chronic disease progression or long-term customer behavior. Finally, pre-training objectives inherited from NLP (e.g., masked token prediction) may need to be adapted to better reflect downstream ES tasks, which often focus on risk scores, anomaly detection, or recommendation quality rather than next-token accuracy.

These challenges are compounded for multimodal foundation models, which remain largely unexplored for ES. The multimodal contrastive methods reviewed in Section~\ref{subsubsec:multimodal-contrastive} align inherently multimodal event data, such as clinical measurements and notes, but only within single domains and at moderate scale. A foundation-scale counterpart that jointly embeds structured event attributes and unstructured content across domains does not yet exist, though the event-level alignment objectives already developed offer natural building blocks.

At present, we are not aware of a widely adopted, domain-agnostic ``foundation model for event streams'' that plays a similar role to GPT-style models in NLP. Existing ES methods, including large-scale masked or autoregressive models in recommender systems and healthcare, still tend to be developed and evaluated within a single domain, on proprietary data, and with task-specific objectives. On the infrastructure side, tools such as Event Stream GPT (ESGPT)~\cite{mcdermott2023event} have begun to provide open-source pipelines for building LLM-style models over continuous-time event sequences, offering unified data preprocessing and standardized evaluation procedures for few-shot and zero-shot tasks. While such toolkits represent valuable progress, they remain focused on specific domains (EHR data in the case of ESGPT), and their broader adoption has been limited. In our view, moving towards true foundation models for ES will require (1) unified representations that can accommodate heterogeneous event attributes and multi-entity structure, (2) large, diverse, and privacy-preserving pre-training corpora, and (3) pretext objectives that jointly capture timing, content, and relational patterns. Advancing along these axes would connect the domain-agnostic SSL vision outlined in Section~\ref{subsec:domain-agnostic-learning} with the broader foundation model agenda for temporal and event-based data. Such progress would also help bridge the gap between ES-specific methods and the rapidly maturing ecosystem of foundation models for related modalities such as time series~\cite{liang2024foundation}, enabling cross-pollination of techniques and shared evaluation protocols.

\section{Conclusion and Discussion}
\label{sec:conclusion}

SSL has emerged as a promising approach to address the inherent challenges of modeling ES data across diverse domains, including healthcare, e-commerce, finance, and gaming. Despite its potential, the field remains fragmented, with predictive SSL paradigms dominating research efforts and other methodologies, such as contrastive learning, underexplored. The lack of unified frameworks, standardized data formats, open benchmarks, and comprehensive datasets further constrains progress.



While the scarcity of contrastive modeling methods in the literature may be potentially explained by a bias against publishing negative results, both in general machine learning research and domain-specific fields, we posit that recent advancements in contrastive learning across various modalities demonstrate its capability to capture meaningful representations and highlight its promise for ES modeling. Similarly, the development of comprehensive datasets and benchmarks would not only facilitate the evaluation of these methods but also provide the foundation for more systematic progress in SSL for ES.


To advance the field, future research should focus on developing domain-agnostic ES models, systematically integrating event timing, and exploring innovative paradigms such as contrastive, joint-embedding predictive, and hybrid architectures. These steps are vital for achieving scalable and generalized SSL solutions capable of unlocking the full potential of ES data for downstream tasks.


Broadly, the maturation of SSL for ES data has the potential to transform industries by enabling robust, scalable models that can operate with minimal labeled data, reducing dependency on costly and impractical labeling efforts. In healthcare, such advances could lead to more accurate patient risk predictions and personalized treatments; in e-commerce, they could refine recommender systems to enhance user experiences. However, these innovations must be developed responsibly, with attention to data privacy, fairness, and transparency to ensure equitable and ethical deployment across applications, particularly in domains where event logs are sensitive or operationally critical.

\begin{acks}
The authors would like to express their gratitude to Alexandra Stark, Tim Elgar, Sahar Asadi, and Bj\"orn Brinne for their invaluable support at various stages of this work. We also extend our sincere thanks to Gabriela Zarzar Gandler and Filip Cornell for their helpful comments and insightful discussions, which greatly contributed to refining and strengthening this research. This work was partially funded by Wallenberg AI, Autonomous Systems and Software Program (WASP).
\end{acks}

\printbibliography

\appendix
\section{Self-Supervised Learning Paradigms}
\label{appendix:ssl-paradigms}

This appendix formalizes the SSL objectives reviewed in Section~\ref{sec:ssl-es-review}, using the notation introduced in Section~\ref{subsec:def-notation}.

\subsection{Predictive SSL}

Predictive objectives train a model to reconstruct or forecast parts of an event stream from the remaining context.

\subsubsection{Masked Modeling}
\label{appendix:masked-modeling}

The masked objective for ES is defined as follows:
given an event stream $S_u$, first a set of indices is sampled $\mathcal{I} \subseteq \{i\}_{i=1}^{n_u}$, then a perturbed version of $S_u$ is constructed by defining its elements as:

$$
\Tilde{e}_i =
\begin{cases} 
\texttt{[MASK]}, & \text{if } i \in \mathcal{I}, \\
e_i, & \text{otherwise}
\end{cases}
\quad \text{for } i=1,\ldots,n_u,
\qquad
\Tilde{S}_u = \{\Tilde{e}_i\}_{i=1}^{n_u}.
$$

This masked objective is parametrized by a neural network $f_\theta$ predicting the masked events, and it can be defined as follows:

$$\mathcal{L}_M = \sum_{i \in \mathcal{I}} - \log p_\theta (e_i | \Tilde{S}_u)$$

Since each event $e_i$ consists of two elements: a timestamp $t_i$ and the event data $d_i$, we can split the above objective into two terms:

$$\mathcal{L}^t_M = \sum_{i \in \mathcal{I}} - \log p_\theta (t_i | \Tilde{S}_u) \quad \text{and} \quad \mathcal{L}^d_M = \sum_{i \in \mathcal{I}} - \log p_\theta (d_i | \Tilde{S}_u),$$

for the event timestamp and data respectively.
If the timestamp does not have to be modeled, the masking process can be altered such that only the event information is perturbed: $\Tilde{e}_i = (t_i, \texttt{[MASK]})$.
In this case $\mathcal{L}_M = \mathcal{L}^d_M$.

In practice, popular MLM models such as BERT~\cite{devlin2018bert} incorporate additional strategies beyond simple masking to enhance model robustness. Specifically, a small subset of the indices selected in $\mathcal{I}$ is replaced with randomly sampled tokens instead of being masked. While this variation introduces minor adjustments to the procedure, it does not fundamentally alter the underlying optimization objective.

\subsubsection{Autoregressive Modeling}
\label{appendix:autoregressive-modeling}

In AR modeling, the objective is to predict the next event in a sequence based on its preceding events, thereby enforcing a causal structure.
For an event stream $ S_u = \{ e_1, e_2, \dots \} $, the model is restricted to using only past events up to $ e_{i-1} $ when predicting $ e_i $, ensuring it learns the temporal dependencies in the sequence.

Given a neural network $ f_\theta $ parameterized by $\theta$, the AR objective can be defined as maximizing the conditional probability of each event given its history. 
This results in the following loss function:

$$
\mathcal{L}_{AR} = - \sum_{i=1}^{n_u} \log p_\theta(e_i | e_{<i})
$$

where \( e_{<i} = \{ e_1, e_2, \dots, e_{i-1} \} \) denotes the history of events up to \( e_{i-1} \). 
Similarly to the masked objective, this formulation can also be split into two by considering the prediction of event data and timestamp separately.

\subsubsection{Temporal Point Processes}
\label{appendix:tpp-formalism}

The key quantity to describe a point process with is the conditional intensity function $\lambda(t \mid \mathcal{H}_t)$, which represents the instantaneous rate of events occurring at time $t$, given the past history $\mathcal{H}_t$. 
For a marked point process, we define the history as:

$$ \mathcal{H}_t = \{(t_i, m_i) : t_i < t \text{ and } m_i \in \mathcal{M} \}, $$

and the intensity function conditioned on the history as

$$\lambda(t, m \mid \mathcal{H}_t) = \lambda(t \mid \mathcal{H}_t) p(m \mid t, \mathcal{H}_t)$$

where $\lambda(t \mid \mathcal{H}_t)$ is the intensity of an event at time $t$ and $p(m \mid t, \mathcal{H}_t)$ is the conditional probability of the mark $m$ given the event time and history.

Similarly to the masked and autoregressive objectives, the objective function for TPPs can be divided into two parts: the likelihood of event times and the likelihood of marks. 
The overall loss function used to optimize the parameters of a marked TPP is defined as
$$
\mathcal{L}_{\text{TPP}} = - \sum_{i=1}^{n} \left( \log \lambda^*(t_i) - \int_{t_{i-1}}^{t_i} \lambda^*(\tau) d\tau + \log p(m_i \mid t_i, \mathcal{H}_{t_i}) \right)
$$

In this formulation, the first term, $\log \lambda^*(t_i)$ represents the log-likelihood of observing an event at time $t_i$, based on the conditional intensity function. For the sake of simplicity we denote $\lambda(t \mid \mathcal{H}_t)$ as $\lambda^*(t)$. The second term, $-\int_{t_{i-1}}^{t_i} \lambda^*(\tau) d\tau$ accounts for the fact that no events occurred between $t_{i-1}$ and $t_i$ and captures the likelihood of no event occurring in this interval (i.e., the survival function). Finally, the third term, $\log p(m_i \mid t_i, \mathcal{H}_{t_i})$ is the log-likelihood of observing the mark $m_i$, given the event history up to time $t_i$.

\subsection{Contrastive SSL}

Contrastive objectives learn representations by aligning augmented views of the same event stream, with or without explicit negative samples.

\subsubsection{Instance Contrastive}
\label{appendix:instance-contrast}

In instance-based contrasting, a positive sample is constructed by applying a transformation $a \sim \mathcal{A}$ to an event stream $S_u$, yielding an augmented view $\Tilde{S}_u$ that preserves semantic similarity. Negative samples are drawn from a set of other sequences, denoted $\mathcal{S}^-$. A neural network $f_\theta$ is then used to encode sequences into their latent representations, with the representation of $S_u$ expressed as $f_\theta(S_u) = z_u$.

A common learning objective for instance-based contrastive learning is the InfoNCE loss~\cite{oord2018representation}, defined as:

$$\mathcal{L}_{ICL} = - \log \frac{\exp(\text{sim}(z_u, \Tilde{z}_u))}{\exp(\text{sim}(z_u, \Tilde{z}_u)) + \sum\limits_{S_j \in \mathcal{S}^-} \exp(\text{sim}(z_u, z_j))},$$

where $\text{sim}(\cdot, \cdot)$ is a similarity metric such as cosine similarity.

\subsubsection{Distillation}
\label{appendix:distillation}

In distillation-based learning, the objective is to minimize the distance between the student's and teacher's latent representations.
Two networks with identical architecture but distinct parameters are used, where the student network is denoted as $f_\theta$ and the teacher network as $f_\zeta$. 
Both networks receive different augmented views of the same input sequence $S_u$, denoted $\Tilde{S}_u$ for the student and $\hat{S}_u$ for the teacher. These views are encoded into latent representations, $\Tilde{z}_u$ for the student and $\hat{z}_u$ for the teacher.

Formally, the student’s encoded view is $f_\theta(\Tilde{S}_u) = \Tilde{z}_u$, and the teacher’s view is $f_\zeta(\hat{S}_u) = \hat{z}_u$. The student network also includes a projection head $q_\theta$.

The distillation loss is defined as:

$$ 
\mathcal{L}_{DL} = - \frac{\langle q_\theta(\Tilde{z}_u), \hat{z}_u \rangle}{\parallel q_\theta(\Tilde{z}_u) \parallel_2 \parallel \hat{z}_u \parallel_2}
$$

where $q_\theta(\Tilde{z}_u)$ is the student's projected latent view, and the loss measures the alignment between the student and teacher representations while maintaining stability in training.

\subsubsection{Feature Decorrelation}
\label{appendix:feature-decor}

Feature Decorrelation-based learning seeks to enforce representation learning by minimizing redundancy between the learned features, thereby avoiding the need for negative samples. 
In contrast to instance-based or distillation-based methods, this approach promotes diversity in the learned representations by directly encouraging feature decorrelation across dimensions.

In this paradigm, two differently augmented views of the same input sequence $S_u$ are generated, denoted $\Tilde{S}_u$ and $\hat{S}_u$, and are passed through a shared neural network $f_\theta$ to produce latent representations $\Tilde{z}_u = f_\theta(\Tilde{S}_u)$ and $\hat{z}_u = f_\theta(\hat{S}_u)$.
To achieve feature decorrelation, two objectives are typically combined: a similarity objective to ensure that the representations of the augmented views are aligned, and a redundancy reduction objective to ensure the latent features are uncorrelated across dimensions.
A loss function for this objective, borrowed from~\cite{zbontar2021barlow}, is as follows:

$$\mathcal{L}_{FDL} = \sum\limits_i(1-C_{ii})^2 + \lambda \sum\limits_{i \neq j} C^2_{ij},$$

where $C$ is the cross-correlation matrix between the features $\Tilde{z}_u$ and $\hat{z}_u$, and $\lambda$ is a weighting factor to balance the two terms. This loss encourages both similarity in aligned representations and decorrelation across feature dimensions.

\subsubsection{Multimodal Contrastive}
\label{appendix:multimodal-learning}

Multimodal learning focuses on leveraging information from different data modalities to capture richer, more comprehensive representations. 
The key assumption is that we have aligned data from two modalities, for example image-text pairs, as it was originally introduced in CLIP~\cite{radford2021learning}.
Due to the need for multiple data modalities, for ES data we only consider multimodal contrasting at the event level (i.e. contrasting between events).

Assuming that the event data domain $\mathcal{D}$ consists of two modality-specific component domains
$\mathcal{D}_a$ and $\mathcal{D}_b$ such that
$$\mathcal{D} = \mathcal{D}_a \times \mathcal{D}_b,$$
and we have sequences $S_u$ where for each event the event data $d_i \in \mathcal{D}$ can be split into two parts 

$$d_i = (d_i^a, d_i^b) \quad \text{where} \quad d_i^a \in \mathcal{D}_a \wedge d_i^b \in \mathcal{D}_b$$

First, a batch of events $\mathcal{B}$ is sampled, consisting of paired instances from the two modalities, such that:

$$
\mathcal{B} = \{(t_i, (d^a_i, d^b_i))\}_{i=1}^{N_b}
$$

Latent representations for each data modality are obtained through their respective encoders $f_\theta^a(d_i^a) = z_i^a$ and $ f_\phi^b(d_i^b) = z_i^b$.
The objective in Multimodal Learning is to align the representations of these paired instances while also maintaining distinctions between unrelated pairs.

The learning objective is typically based on the InfoNCE loss, adjusted to handle the multimodal nature of the data. 
The specific loss function for multimodal learning (denoted $\mathcal{L}_{MML}$) is:

\begin{align*}
\mathcal{L}_{MML} &= \frac{1}{2N_b} \left( \sum_{i=1}^{N_b} \left( - \log \frac{\exp(\text{sim}(z^a_i, z^b_i))}{\sum_{j} \exp(\text{sim}(z^a_i, z^b_j))} \right) \right. \\
&\quad + \left. \sum_{i=1}^{N_b} \left( - \log \frac{\exp(\text{sim}(z^b_i, z^a_i))}{\sum_{j} \exp(\text{sim}(z^b_i, z^a_j))} \right) \right)
\end{align*}

where $N_b$ is the batch size and $\text{sim}(\cdot, \cdot)$ refers to a similarity metric such as cosine similarity.
This loss encourages the representations from both modalities to be well-aligned while simultaneously discouraging similarities between non-matching pairs.
Note that this formulation (and multimodal learning in general) is applied more as an event-encoder rather than a full sequence encoder, unlike other contrastive learning paradigms presented here.

\section{Reproducibility Checklist for JAIR}

\subsection*{All articles:}

\begin{enumerate}
    \item All claims investigated in this work are clearly stated.
    [yes]
    \item Clear explanations are given how the work reported substantiates the claims.
    [yes]
    \item Limitations or technical assumptions are stated clearly and explicitly.
    [yes]
    \item Conceptual outlines and/or pseudo-code descriptions of the AI methods introduced in this work are provided, and important implementation details are discussed.
    [NA]
    \item Motivation is provided for all design choices, including algorithms, implementation choices, parameters, data sets and experimental protocols beyond metrics.
    [yes]
\end{enumerate}

\subsection*{Articles containing theoretical contributions:}
Does this paper make theoretical contributions?
[no]

\subsection*{Articles reporting on computational experiments:}
Does this paper include computational experiments?
[no]

\subsection*{Articles using data sets:}
Does this work rely on one or more data sets (possibly obtained from a benchmark generator or similar software artifact)?
[no]

\subsection*{Explanations on any of the answers above (optional):}

This article is a survey: it introduces no new algorithms, experiments, or data sets. Publicly available data sets discussed in the survey are cited in Section~\ref{subsec:datasets-overview}.

\end{document}